\documentclass[acmlarge,screen,nonacm]{acmart}

\usepackage{graphicx}
\usepackage{longtable}
\usepackage{multirow,colortbl}
\usepackage{titlesec}
\usepackage{url}
\usepackage{adjustbox}

\newcommand{\tmpframe}[1]{\fbox{#1}}
\def\BibTeX{{\rm B\kern-.05em{\sc i\kern-.025em b}\kern-.08emT\kern-.1667em\lower.7ex\hbox{E}\kern-.125emX}}

\begin{document}

%
% The "title" command has an optional parameter, allowing the author to define a "short title" to be used in page headers.
\title{Recent Advances and Trends in Multimodal Deep Learning: A Review}

%
% The "author" command and its associated commands are used to define the authors and their affiliations.
% Of note is the shared affiliation of the first two authors, and the "authornote" and "authornotemark" commands
% used to denote shared contribution to the research.
\author{Jabeen Summaira}
\email{11821129@zju.edu.cn}
\author{Xi Li}
\authornote{Corresponding Author: xilizju@zju.edu.cn}
\email{xilizju@zju.edu.cn}
\affiliation{%
	\institution{College of Computer Science, Zhejiang University, China}
	\streetaddress{P.O. Box W-99}
	\city{Hangzhou}
	\postcode{310027}
}

\author{Amin ~Muhammad Shoib}
\affiliation{%
	\institution{School of Software Engineering, East China Normal University}
	\streetaddress{3663 North Zhongshan Road.}
	\city{Shanghai}
	\country{China}}
\email{52184501030@stu.ecnu.edu.cn}

\author{Omar Bourahla}
\email{bourahla@zju.edu.cn}

\author{Li Songyuan}
\email{leizungjyun@zju.edu.cn}

\author{Jabbar Abdul}
\email{Jabbar@zju.edu.cn}
\affiliation{%
	\institution{College of Computer Science, Zhejiang University, China}
	\streetaddress{P.O. Box W-99}
	\city{Hangzhou}
	\postcode{310027}
}
%
% By default, the full list of authors will be used in the page headers. Often, this list is too long, and will overlap
% other information printed in the page headers. This command allows the author to define a more concise list
% of authors' names for this purpose.
\renewcommand{\shortauthors}{J. Summaira, et al.}

%
% The abstract is a short summary of the work to be presented in the article.
\begin{abstract}
Deep Learning has implemented a wide range of applications and has become increasingly popular in recent years. The goal of multimodal deep learning is to create models that can process and link information using various modalities. Despite the extensive development made for unimodal learning, it still cannot cover all the aspects of human learning. Multimodal learning helps to understand and analyze better when various senses are engaged in the processing of information. This paper focuses on multiple types of modalities, i.e., image, video, text, audio, body gestures, facial expressions, and physiological signals. Detailed analysis of past and current baseline approaches and an in-depth study of recent advancements in multimodal deep learning applications has been provided. A fine-grained taxonomy of various multimodal deep learning applications is proposed, elaborating on different applications in more depth. Architectures and datasets used in these applications are also discussed, along with their evaluation metrics. Last, main issues are highlighted separately for each domain along with their possible future research directions. 

\end{abstract}



%
% Keywords. The author(s) should pick words that accurately describe the work being
% presented. Separate the keywords with commas.

%
% A "teaser" image appears between the author and affiliation information and the body 
% of the document, and typically spans the page. 
%%  \includegraphics[width=\textwidth]{sampleteaser}
%%  \caption{Seattle Mariners at Spring Training, 2010.}
%%  \Description{Enjoying the baseball game from the third-base seats. Ichiro Suzuki preparing to bat.}

%
% This command processes the author and affiliation and title information and builds
% the first part of the formatted document.
\maketitle

\section{Introduction:}
Machine learning (ML) has become increasingly popular in recent studies. It has been implemented in a vast application range such as image recognition, multimedia concept retrieval, social network analysis, video recommendation, text mining, etc. Deep Learning (DL) is widely used in these applications \cite{pouyanfar2018survey}. The exponential growth, incredible developments, and data availability in computing technologies have contributed to the rise of DL research. DL success has been a motivating factor for solving even more complex ML problems. Additionally, DL's principal advantage is its representation in hierarchical form, i.e., it can learn effectively through a general-purpose learning process. Various new DL approaches have been developed and have shown impressive outcomes across multiple applications such as visual data processing, Natural Language Processing (NLP), voice and audio processing, and many other widely-known applications. Multimodal Deep learning (MMDL) has recently attracted significant research intentions due to the development of DL. A list of abbreviations is presented in the Table \ref{tab:Abb_Long}.

Our experience of things around us is multimodal; we see things, hear, touch, smell, and taste. Multiple aspects of the object are captured to transmit information in different media forms like image, text, video, graph, sound, etc. Modality specifies a representation format in which a particular type of information is stored. Therefore, various media forms mentioned above relate to modalities, and the representation of these multiple modalities together can be defined as multimodal \cite{guo2019deep}. Extensive development has been made in terms of uni-modal. Still, it is insufficient to model the complete aspects of humans. Uni-modal works better where progress of the method is required in only one mode. Multimodal learning suggests we understand and analyze better when a variety of senses are engaged in the processing of information. This paper focuses on various types of modalities, i.e., image, video, text, audio, body gestures, facial expressions, and physiological signals from the MMDL perspective. The main goal of MMDL is to construct a model that can process information from different modalities and relate it.

\begin{center}	
	\scriptsize 
	\centering
	\begin{longtable}{|p{1.2cm}|p{6cm}|p{1.2cm}|p{6cm}|}
		
		\caption{List of Abbreviations}
		\label{tab:Abb_Long} \\
		
		\bottomrule
		\multicolumn{1}{|c|}{\cellcolor{blue!20}\textbf{Abbreviation}} & \multicolumn{1}{|c|}{\cellcolor{blue!20}\textbf{Explanation}} &
		\multicolumn{1}{|c}{\cellcolor{blue!20}\textbf{Abbreviation}} & \multicolumn{1}{|c|}{\cellcolor{blue!20}\textbf{Explanation}} \\ 
		\hline 
		\endfirsthead
		
		AFDBN	& Adaptive Fractional Deep Belief Network &
		AI		& Artificial Intelligence \\
		AMT		& Amazon Mechanical Turk &
		ANN  	& Artificial Neural Network \\
		ASKMS 	& Adaptive Shape Kernel Based Mean Shift &
		ATTS	& Articulatory Text To Speech \\
		AUC		& Area Under Curve &
		AVSR	& Audio-Visual Speech Recognition\\
		BiLSTM  & Bidirectional Long Short-Term Memory &
		BNDBM	& Batch Normalized Deep Boltzmann Machine \\
		BRNN 	& Bidirectional Recurrent Neural Network&
		C3D		& Three Dimensional Convolutional Neural Networks \\
		CALO-MA & CALO Meeting Assistant &
		CDBN	& Competitive Deep Belief Network \\
		CHIL	& Computers in the Human Interaction Loop &
		CLEVR	& Compositional Language and Elementary Visual Reasoning diagnostics\\
		CMN		& Compositional Modular Network &
		CNN		& Convolutional Neural Network \\
		COH		& Color Oriented Histogram &
		CRF		& Conditional Random Field \\
		CRNN	& Cascade Recurrent Neural Network &
		CssCDBM & Contractive slab and spike Convolutional Deep Boltzmann Machine \\
		CTTS	& Concatenative Text To Speech &
		CV		& Computer Vision \\
		DAQUAR  & DAtaset for QUestion Answering on Real-world images&
		DBM 	& Deep Boltzmann Machine \\
		DBN		& Deep Belief Network &
		DDBN	& Discriminative Deep Belief Network \\
		DL		& Deep Learning &
		DLTTS	& Deep Learning Text To Speech \\
		DNN		& Deep Neural Network &
		DRL		& Deep Reinforcement Learning \\
		E2E		& End-to-End &
		EDR		& Equal Detected Rate \\
		EER		& Equal Error Rate &
		EOH 	& Edge Oriented Histogram \\
		F3RBM	& Fuzzy Removing Redundancy Restricted Boltzmann 	Machine &
		FDBN	& Fuzzy Deep Belief Network \\
		FRBM	& Fuzzy Restricted Boltzmann Machine &
		FTTS	& Formant Text To Speech \\
		FVQA	& Fact-based Visual Question Answering &
		GBRBM	& Gaussian-Bernoulli Restricted Boltzmann Machine \\
		GLU		& Gated Linear Unit &
		GMM		& Gaussian Mixture Model \\
		GRU		& Gated Recurrent Unit &
		HMM		& Hidden Markov Model \\
		HRL 	& Hierarchical Reinforcement Learning &
		IEMOCAP	& Interactive Emotional Dyadic Motion Capture\\
		KB		& Knowledge Bases &
		LSTM	& Long Short-Term Memory \\
		MAE		& Mean Absolute Error &
		mAP		& mean Average Precision \\
		MCB		& Multimodal Compact Bilinear &
		MCD		& Mel Cepstral Distortion\\
		MDP		& Markov Decision Process  &
		MFCC	& Mel Frequency Cepstral Coefficients\\
		ML	 	& Machine Learning &
		MMAM	& Multimodal Attention-Based Models \\
		MMDL	& Multimodal Deep Learning&
		MMED	& Multimodal Event Detection\\
		MMER	& Multimodal Emotion Recognition &
		MMEKBM	& Multimodal External Knowledge Bases Models\\
		MMJEM	& Multimodal Joint-Embedding Models &
		MOS		& Mean Opinion Score \\
		MS-COCO	& Microsoft-Common Objects in Context&
		msDBM 	& Mean supervised Deep Boltzmann Machine\\
		MSR-VTT	& Microsoft Research-Video to Text&
		MSVD	& Microsoft Video Description \\
		MuSE 	& Multimodal Stressed Emotion &
		M-VAD	& Montreal-Video Annotation Dataset\\
		NLP		& Natural Language Processing&
		NMI		& Normalized Mutual Information \\
		NN		& Neural Networks &
		NRBM	& Normalized Restricted Boltzmann Machine \\
		PoS		& Part of Speech &
		PPLX	& Perplexity metric \\
		PTTS	& Parametric Text to Speech &
		RBM  	& Restricted Boltzmann Machine \\
		RBS		& Rule Based System &
		RECOLA  & REmote COLlaborative and Affective \\
		RF\_TF	& Random-Forest Tag-Propagation &
		RGB-D	& Red Green Blue-Depth \\
		RNN		& Recurrent Neural Network &
		RossDBM & Robust spike-and-slab Deep Boltzmann Machine \\
		SEMAINE & Sustained Emotionally coloured Machinehuman Interaction using Nonverbal Expression &
		SGRU	& Stacked Gated Recurrent Unit\\
		SiGRU	& Simplified GRU &
		SML		& Stochastic Maximum Likelihood \\
		SMT		& Statistical Machine Translation &
		SRE 	& Semantic Relation Extractor \\
		SST		& Single-Stream Temporal &
		STFT	& Short-Time Fourier Transform \\
		SVO		& Subject, Object, Verb &
		TDIUC	& Task Directed Image Understanding Challenge\\
		TTS		& Text To Speech &
		V2C		& Video-to-Commonsense \\
		VAE		& Variational Auto-Encoders &
		VCTK	& Voice Cloning Toolkit \\
		VDR		& Visual Dependency Representation&
		VQA		& Visual Question-Answering  \\
		\bottomrule
	\end{longtable}
\end{center}

The future of Artificial Intelligence (AI) has been revolutionized by DL. It has addressed several complex problems that have existed for many years within the AI community. For MMDL, various deep architectures with different learning frameworks are rapidly designed. Machines are developed to perform similarly to humans or even better in other application areas such as self-driving cars, image processing, medical diagnosis and predictive forecasting etc \cite{sengupta2020review}. The recent advances and trends of MMDL are from Audio-visual speech recognition (AVSR) \cite{yuhas1989integration}, multimodal emotion recognition \cite{chen2021heu}, image and video captioning \cite{liu2021cptr,hosseinzadeh2021video}, Visual Question-Answering (VQA) \cite{xi2020visual}, to multimedia retrieval \cite{souza2021online} and so on.

\subsection{Contribution and Relevance to other surveys:}

Recently , numerous surveys (such as, C. Zhang et al. \cite{zhang2020multimodal} 2020, Y. Bisk et al. \cite{bisk2020experience} 2020, J. Gao et al. \cite{gao2020survey} 2020, A. Mogadala et al. \cite{mogadala2019trends} 2019, SF Zhang et al. \cite{zhang2019multimodal} 2019, W Guo et al. \cite{guo2019deep} 2019, T Baltrušaitis et al. \cite{baltruvsaitis2018multimodal} 2018, Y Li et al. \cite{li2018survey} 2018, and D Ramachandram et al. \cite{ramachandram2017deep} 2017 are already published relating to the topic of multimodal learning. These review articles are analyzed and discussed further in the section \ref{AnalysisBaselineSurveys}. Our contributions to this article are described in section \ref{contribution}. A summarized list of these review articles is presented in Table \ref{tab:relatedSurveys}. 

\subsubsection{Analysis of baseline surveys:}
\label{AnalysisBaselineSurveys}

C. Zhang et al. \cite{zhang2020multimodal} discussed natural language and vision modalities in NLP and Computer Vision (CV). 
Y. Bisk et al. \cite{bisk2020experience} reviewed work based on NLP contextual foundations.  
The fundamentals of MMDL fusion methods and their challenges are discussed by Gao et al. \cite{gao2020survey}.
Mogadala et al. \cite{mogadala2019trends} mentioned ten different groups of approaches based on language and vision tasks. Datasets, evaluation metrics used in these approaches, and their results are also discussed. 
SF Zhang et al. \cite{zhang2019multimodal}, W Guo et al. \cite{guo2019deep}, and Y Li et al. \cite{li2018survey} explained different algorithms and applications of multimodal representation learning, its recent trends and challenges.
T Baltrušaitis et al. \cite{baltruvsaitis2018multimodal} described various frameworks based on five core challenges, i.e., representation, translation, alignment, fusion and co-learning of multimodal machine learning.  
D Ramachandram et al. \cite{ramachandram2017deep} examined recent developments in deep multimodal learning and limitations and obstacles in this active area of study; also regularization strategies and methods to optimize the structure of multimodal fusion are highlighted.  
\subsubsection{Our contributions:}
\label{contribution}
All these literature surveys provide a review on only a specific domain of multimodal learning. Some authors only discussed methods and applications of representation learning and some on fusion learning. Others discussed representation and fusion learning methods together using few modalities like vision or text. As mentioned earlier, Mogadala et al. \cite{mogadala2019trends} explained different models of MMDL using image, video and text modalities. However, we have taken one step further; along with these modalities, this paper provides a technical review of various models using audio, body gestures, facial expressions and physiological signals modalities. Our primary focus is distinctive in that we seek to survey the literature from up-to-date deep learning concepts using more modalities. The main contributions of our article are listed below:
\begin{itemize}
	\item A novel fine-grained taxonomy of various Multimodal Deep Learning applications is proposed, which elaborates different groups of applications in more depth.
	\item Various models using the image, video, text, audio, body gestures, facial expressions and physiological signals modalities are discussed together.
	\item Recent advancements of MMDL in image description is discussed in section \ref{EDID}, \ref{SCID} and \ref{AID}.
	\item In video description and VQA tasks, up-to-date literature is discussed in section \ref{CNN-RNN VidD}, \ref{Rnn-Rnn VidD}, \ref{DRL VidD} and \ref{VQAJoint}, \ref{VQAAttention}, \ref{VQAknowledge} respectively. 
	\item In Section \ref{DLTTS} and \ref{EmotionRecog}, the contribution of deep learning frameworks about speech synthesis and emotion recognition is explained in detail. 
	\item Main issues are highlighted separately for each application group, along with their possible future research directions.   
	
\end{itemize}

\begin{table}
	\caption{List of related literature surveys}
	\label{tab:relatedSurveys}
	\small
	\begin{tabular}{p{0.3cm}p{13cm}p{1cm}p{0.6cm}}
		\bottomrule
		\cellcolor{blue!20}\textbf{No.}  & \cellcolor{blue!20}\textbf{Title} &  \cellcolor{blue!20}\textbf{Pub.}  & \cellcolor{blue!20}\textbf{Year} \\
		\bottomrule
		1 & Multimodal Intelligence: Representation Learning, Information Fusion, and Applications \cite{zhang2020multimodal} & IEEE & 2020 \\
		2 & Experience Grounds Language \cite{bisk2020experience} & arXiv & 2020 \\
		3 & A Survey on Deep Learning for Multimodal Data Fusion \cite{gao2020survey} & NC(MIT) & 2020 \\
		4 & Trends in integration of vision and language research: A survey of tasks, datasets, and methods \cite{mogadala2019trends} & arXiv & 2019 \\
		5 & Multimodal Representation Learning: Advances, Trends and Challenges \cite{zhang2019multimodal} & IEEE & 2019 \\
		6 & Deep multimodal representation learning: A survey \cite{guo2019deep} & IEEE & 2019 \\
		7 & Multimodal Machine Learning: A Survey and Taxonomy \cite{baltruvsaitis2018multimodal} & IEEE & 2018 \\
		8 & A Survey of Multi-View Representation Learning \cite{li2018survey} & IEEE & 2018 \\
		9 & Deep Multimodal Learning: A survey on recent advances and trends \cite{ramachandram2017deep} & IEEE & 2017 \\
		\bottomrule
	\end{tabular}
\end{table}

\subsection{Structure of Survey:}

The rest of the article is structured as follows. Some background of MMDL is discussed in section \ref{backGround}. Section \ref{AppsMMDL} presents various MMDL application groups. Architectures used in MMDL are discussed in section \ref{architectMMDL}. Dataset and evaluation metrics are discussed in sections \ref{dataset} and \ref{E-Metrics}, respectively. In section \ref{Disc&future}, we provided a brief discussion and propose future research directions in this active area. Finally, section \ref{conclusion} concludes the survey paper.

\section{Background:} 
\label{backGround}

A lot of ML approaches have been developed over the past decade to deal with multimodal data. Multimodal applications came into existence in the 1970s and were categorized into four eras \cite{Morency}. Description of these eras is explained in section \ref{4eras}. In this paper, we concentrate mainly on deep learning era for multimodal applications. 

\subsection{Eras of multimodal learning:}
\label{4eras}

Prior research on multimodal learning can be categorized into four eras, listed below.

\begin{itemize}
	\item "Behavioral Era (BE)": 
	The BE starts from the 1970s to the late 1980s. During this era, AA Lazarus \cite{lazarus1973multimodal} proposed Multimodal behavior therapy based on seven different personality dimensions using inter-related modalities. 
	RM Mulligan \& ML Shaw \cite{mulligan1980multimodal} presents multimodal signal detection, wherein the signal detection task, the pooling of information is examined from other modalities, i.e., auditory and visual stimuli.

	\item "Computational Era (CE)": 
	The CE starts from the late 1980s to 2000. During this era, ED Petajan \cite{petajan1985automatic} proposed an "Automatic Lipreading to Enhance Speech Recognition." BP. Yuhas et al. \cite{yuhas1989integration} increase the performance of automatic speech recognition system even in noisy environments by using neural networks (NN). Hidden Markov Model (HMM) \cite{juang1991hidden} has become so popular in speech recognition because of the inherent statistical framework, the ease and accessibility of training algorithms to estimate model parameters from finite speech data sets.
	
	\item "Interaction Era (IE)":
	The IE starts from 2000 to 2010. During this era, one of the first milestones is the AMI Meeting Corpus \cite{carletta2005ami} (a multimodal dataset) with over 100 hours of meeting recordings, all thoroughly annotated and transcribed. AMI seeks to create a repository of meetings and to evaluate conversations. CHIL project motive \cite{waibel2009computers} uses ML methods to extract nonverbal human behaviors automatically. CALO Meeting Assistant (CALO-MA) \cite{tur2008calo} real-time meeting recognition system takes the speech models. CALO-MA architecture includes real-time and offline speech transcription, action item recognition, question-answer pair identification, decision extraction, and summarization.
	
	\item "Deep Learning Era (DLE)": 
	The DLE starts in 2010 to date. This deep learning era is the main focus of our review paper. We comprehensively discuss different groups of applications in section \ref{AppsMMDL}, MMDL architectures in section \ref{architectMMDL}, datasets in section \ref{dataset} and evaluation metrics in section \ref{E-Metrics} proposed during this era. 
\end{itemize}

\section{MMDL Applications:}
\label{AppsMMDL}

\begin{figure}
	\centering
	\tmpframe{\includegraphics[width= 130mm, height=70mm]{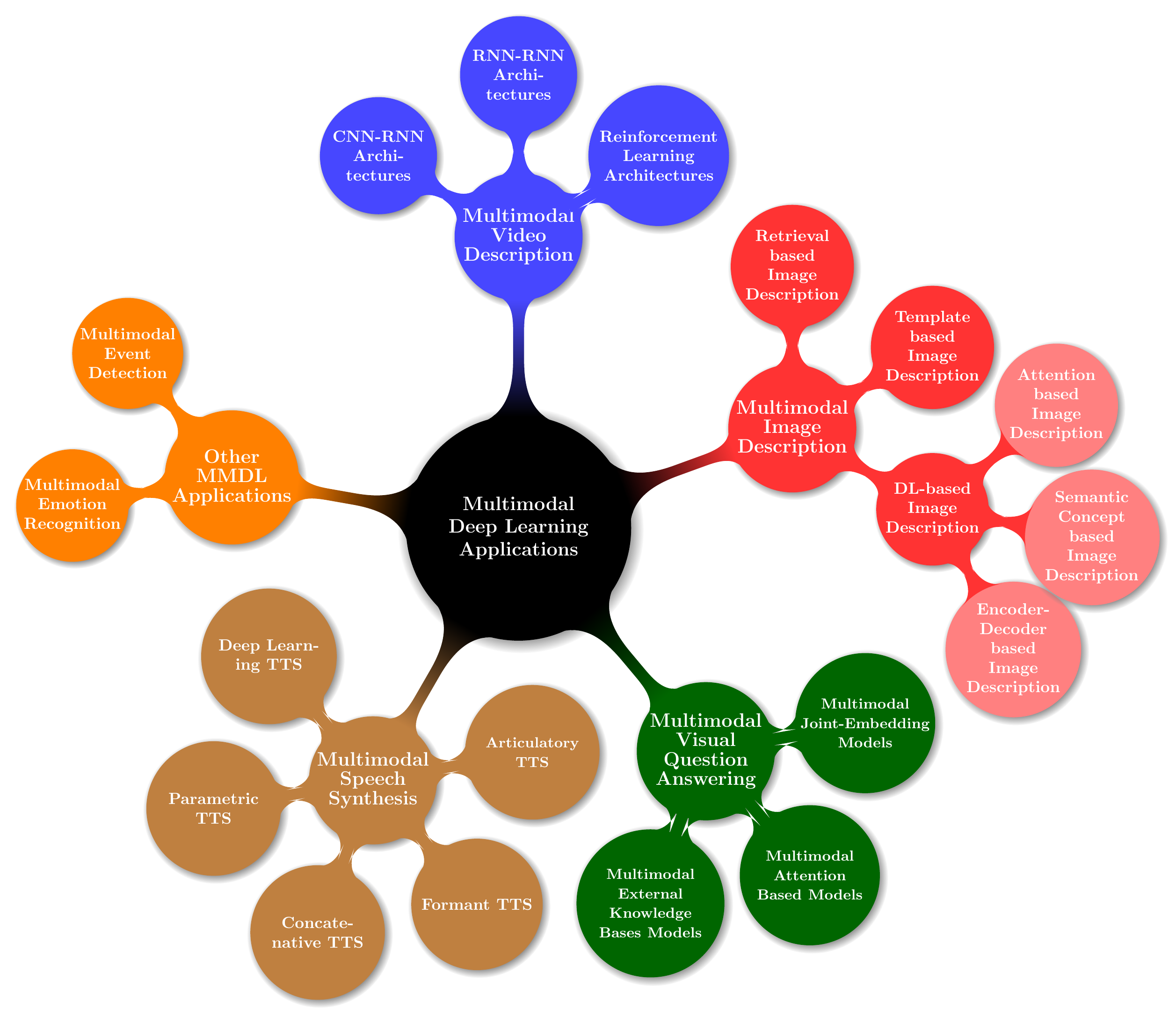}}
	\caption{Taxonomy diagram of Multimodal Deep Learning Applications.}
	\label{fig:appsTexonomy}
\end{figure}

During the DLE, various applications are designed by using multimodal deep learning techniques. In this article, these applications are grouped with relevance and dominance across multiple research areas. The taxonomy diagram of these applications is presented in the Figure \ref{fig:appsTexonomy}.

\subsection{Multimodal Image Description:}
\begin{figure}
	\centering
	\tmpframe{\includegraphics[width=100mm,height=45mm]{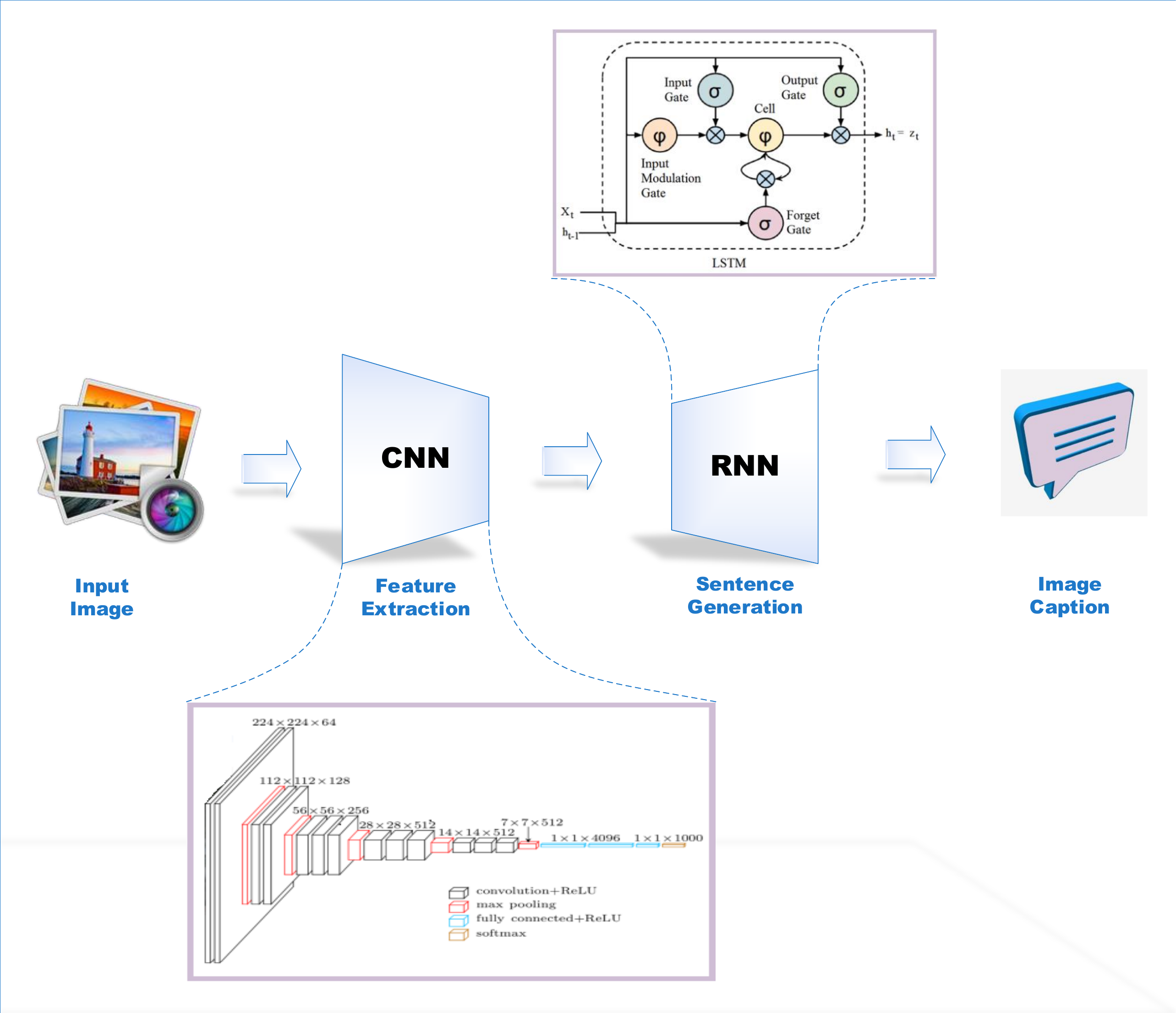}}
	\caption{General structure diagram of Image Description Model.}
	\label{fig:imgDescription}
\end{figure}

Image Description is mostly used to generate a textual description of visual contents provided through input image. During the deep learning era, two different fields are merged to perform image descriptions, i.e., CV and NLP. In this process, two main kinds of modalities are used, i.e., image and text. The image description's general structure diagram is shown in Figure \ref{fig:imgDescription}. Image description frameworks are categorized into Retrieval-based, Template-based, and DL-based image descriptions. Retrieval and Template based image descriptions are one of the earliest techniques for describing visual contents from images. In this article, DL-based image description techniques are explained in detail, which are further categorized into encoder-decoder-based, semantic concept-based, and attention-based image descriptions.

\subsubsection{Encoder-Decoder based Image Description (EDID):}
\label{EDID}
EDID plays a vital role in image captioning tasks using DL architectures. CNN architectures are used mainly as encoder parts to extract and encode data from images, and RNN architectures are used as decoder part to decode and generate captions.
J. Wu \& H.hu \cite{wu2017cascade} proposed a Cascade Recurrent Neural Network (CRNN) for image description. For the learning of visual language interactions, a cascade network is adopted by CRNN from the forward and backward directions. In this approach, two embedding layers are constructed for dense word expression, and Stacked GRU (SGRU) is designed for mappings of the image to its corresponding language model.
R. Hu et al. \cite{hu2017modeling} proposed an end-to-end (E2E) Compositional Modular Network (CMN) model that learns image region localization and language representation jointly. In this model, Referential expressions are observed based on subject-entity, object-entity and a relationship.
L. Guo et al. \cite{guo2019mscap} proposed a multi-style image captioning framework using CNN, GAN, LSTM and GRU architectures. In this framework, five different captioning styles are introduced for the image: romantic, negative, positive, factual and humorous styles.
X. He et al. \cite{he2019image} proposed an image caption generation framework with the guidance of Part of Speech (PoS). In the word generation part, PoS tags are fed into LSTM as guidance to generate more effective image captions. 
Y. Feng et al. \cite{feng2019unsupervised} proposed an unsupervised image captioning framework. In this model first attempt is made to do captioning of the image without any labeled image-sentence pairs. For this purpose, about two million sentences are crawled to facilitate unsupervised scenario of captioning.

\subsubsection{Semantic Concept-based Image Description (SCID):}
\label{SCID}
A collection of semantic concepts extracted from the image are selectively addressed by SCID approaches. These concepts are extracted at the encoding stage along with other features of an image, then merged into hidden states of language models, and output is used to generate descriptions of images based on semantic concepts. 
W. Wang et al. \cite{wang2018novel} proposed an attribute-based image caption generation framework. Visual features are extracted by using salient semantic attributes and are pass as input to LSTM’s encoder. These semantic attributes can extract more semantic related data in images to enhance the generated caption’s accuracy.
P. Cao et al. \cite{cao2019image} proposed a semantic-based model for image description. In this model, semantic attention-based guidance is used for LSTM architecture to produce a description of an image.  
L. Cheng et al. \cite{cheng2020stack} proposed a multi-stage visual semantic attention mechanism based image description model. In this approach, top-down and bottom-up attention modules are combined to control the visual and semantic level information for producing fine-grained descriptions of the image.

\subsubsection{Attention-based Image Description (AID):}
\label{AID}
AID plays a vital role because it helps the image description process by focusing on distinct regions of the image according to their context. In recent years, various techniques have been proposed to better describe an image by applying an attention mechanism. Some of these attention mechanism based image descriptions techniques are; L. Li et al. \cite{li2017gla} proposed a new framework for describing images by using local and global attention mechanisms. Selective object-level features are combined with image-level features according to the context using local and global attention mechanisms. Therefore, more appropriate descriptions of images are generated.  
P. Anderson et al. \cite{anderson2018bottom} proposed bottom-up and top-down attention based framework for image description to encourage deeper image understanding and reasoning. 
M. Liu et al. proposed a dual attention mechanism-based framework to describe an image for Chinese  \cite{liu2020chinese} and English \cite{liu2020image} languages. The textual attention mechanism is used to improve the data credibility, and the visual attention mechanism is used to a deep understanding of image features by integrating various image labels for English and Chinese description generation. 
B. Wang et al. \cite{wang2020cross} proposed an E2E-DL approach for image description using a semantic attention mechanism. In this approach, features are extracted from specific image regions using an attention mechanism for producing corresponding descriptions. This approach can transform English language knowledge representations into the Chinese language to get cross-lingual image description.
Y. Wei et al. \cite{wei2020multi} proposed an image description framework by using multi attention mechanism to extract local and non-local feature representations.    
LU Jiasen et al. \cite{jiasen2020adaptive} proposed an adaptive attention mechanism based image description model. Attention mechanism merges visual features extracted from the image by CNN architecture and linguistic features by LSTM architecture.
\\

During the deep learning era (2010 to date), many authors contributed a lot by proposing various techniques to describe the visual contents of an image in the domain of image description. Different image description approaches described in section \ref{EDID}, \ref{SCID}, and \ref{AID} are analyzed comparatively according to architectures, multimedia, publication year, datasets, and evaluation metrics in Table \ref{tab:ImageDes}. The architectures used in these proposed techniques are explained briefly in section \ref{architectMMDL}. Similarly, datasets and evaluation metrics are discussed in sections \ref{dataset} and \ref{E-Metrics}, respectively.

% \usepackage{multirow}
\begin{table}[]
	\scriptsize 
	\centering
	\caption{Comparative analysis of Image Description models. Where, EDID = Encoder-Decoder based Image Description, SCID = Semantic Concept-based Image Description, and AID =  Attention-based Image Description.}
	\label{tab:ImageDes}
	\begin{adjustbox}{width=1.1\textwidth, center=\textwidth}
		\begin{tabular}{|p{0.4cm}|p{2.4cm}|p{0.2cm}|p{3.2cm}|p{1.3cm}|p{3.6cm}|p{3.3cm}|}
			\bottomrule
			\multicolumn{1}{|l|}{}&\multicolumn{1}{l|}{\textbf{Paper}}&\multicolumn{1}{l|}{\textbf{Year}}&\multicolumn{1}{l|}{\textbf{Architecture}}& \multicolumn{1}{l|}{\textbf{Multimedia}} & \multicolumn{1}{l|}{\textbf{Dataset}}&\multicolumn{1}{l|}{\textbf{Evaluation Metrics}}    \\ 
			\bottomrule

			\multirow{7}{*}{} \cellcolor{orange!25} & J. Wu et al. \cite{wu2017cascade} & 2017 & CNN/VGG16-InceptionV3, Stacked GRU & Image, Text & MS-COCO & BLEU, CIDEr, METEOR \\
			\cellcolor{orange!25}  & R. Hu et al. \cite{hu2017modeling} & 2017 & Faster RCNN/VGG16, RNN/BLSTM & Image, Text & Visual Genome, Google-Ref & Top-1 precision (P@1) metric \\
			\cellcolor{orange!25} EDID& L. Guo et al. \cite{guo2019mscap} & 2019 & Deep CNN, GAN, RNN/LSTM, GRU & Image, Text & FlickrStyle10K, SentiCap, MS-COCO & BLEU, CIDEr, METEOR, PPLX \\ 
			\cellcolor{orange!25}& X. He et al \cite{he2019image} & 2019 & CNN/VGG16, 
			RNN/LSTM & Image, Text & Flickr30k, MS-COCO & BLEU, CIDEr, METEOR\\
			\cellcolor{orange!25}& Y. Feng et al. \cite{feng2019unsupervised} & 2019 & CNN/InceptionV4, RNN/LSTM & Image, Text & MS-COCO & BLEU, ROUGE, CIDEr, METEOR, SPICE \\
			\bottomrule
			
			\multirow{4}{*}{} \cellcolor{red!15} & W. Wang et al. \cite{wang2018novel} & 2018 & CNN/VGG16,
			RNN/LSTM & Image,Text & MS-COCO & BLEU, CIDEr, METEOR\\
			\cellcolor{red!15} SCID & P. Cao et al. \cite{cao2019image} & 2019 & CNN/VGG16, RNN/BLSTM & Image, Text & Flickr8k, MS-COCO& BLEU, CIDEr, METEOR \\
			\cellcolor{red!15} & L. Cheng et al. \cite{cheng2020stack} & 2020 & Faster-RCNN, RNN/LSTM & Image, Text & MS-COCO & BLEU, SPICE, METEOR, CIDEr, ROUGE \\
			\bottomrule
			
			\multirow{8}{*}{} \cellcolor{blue!15} & L. Li et al. \cite{li2017gla} & 2017 & CNN/VGG16-Faster RCNN, RNN/LSTM & Image, Text & Flickr8K, Flickr30K, MS-COCO & METEOR, ROUGE$_{L}$, CIDEr, BLEU \\
			\cellcolor{blue!15}& P. Anderson et al. \cite{anderson2018bottom}  & 2018 & Faster RCNN/ResNet101, RNN/LSTM, GRU & Image, Text & Visual Genome Dataset, MS-COCO, VQA v2.0 & BLEU, METEOR, CIDEr, SPICE, ROUGE \\
			\cellcolor{blue!15}& M. Liu et al. \cite{liu2020chinese} & 2020 & CNN/InceptionV4, RNN/LSTM & Image, Text & Flickr8k-CN, Flickr8k-CN, AIC-ICC & BLEU, ROUGE, CIDEr, METEOR \\
			\cellcolor{blue!15} AID & M. Liu et al. \cite{liu2020image} & 2020 & CNN/InceptionV4, RNN/LSTM & Image, Text & AIC-ICC & BLEU, ROUGE, CIDEr, METEOR \\
			\cellcolor{blue!15}& B. Wang et al. \cite{wang2020cross} & 2020 & CNN/InceptionV4, RNN/LSTM & Image, Text & Flickr8K, Flickr8k-CN & BLEU, ROUGE, CIDEr, METEOR \\
			\cellcolor{blue!15}& Y. Wei et al. \cite{wei2020multi} & 2020 & GAN, RNN/LSTM & Image, Text & MS-COCO & BLEU, METEOR, CIDEr, SPICE, ROUGE \\
			\cellcolor{blue!15}& LU Jiasen et al. \cite{jiasen2020adaptive} & 2020 & CNN/ResNet, RNN/LSTM & Image, Text & Flickr30K, MS-COCO & BLEU, ROUGE, CIDEr, METEOR \\
			\bottomrule
		\end{tabular}
	\end{adjustbox}
\end{table}

\subsection{Multimodal Video Description:}
\begin{figure}
	\centering
	\tmpframe{\includegraphics[width=150mm,height=45
		mm]{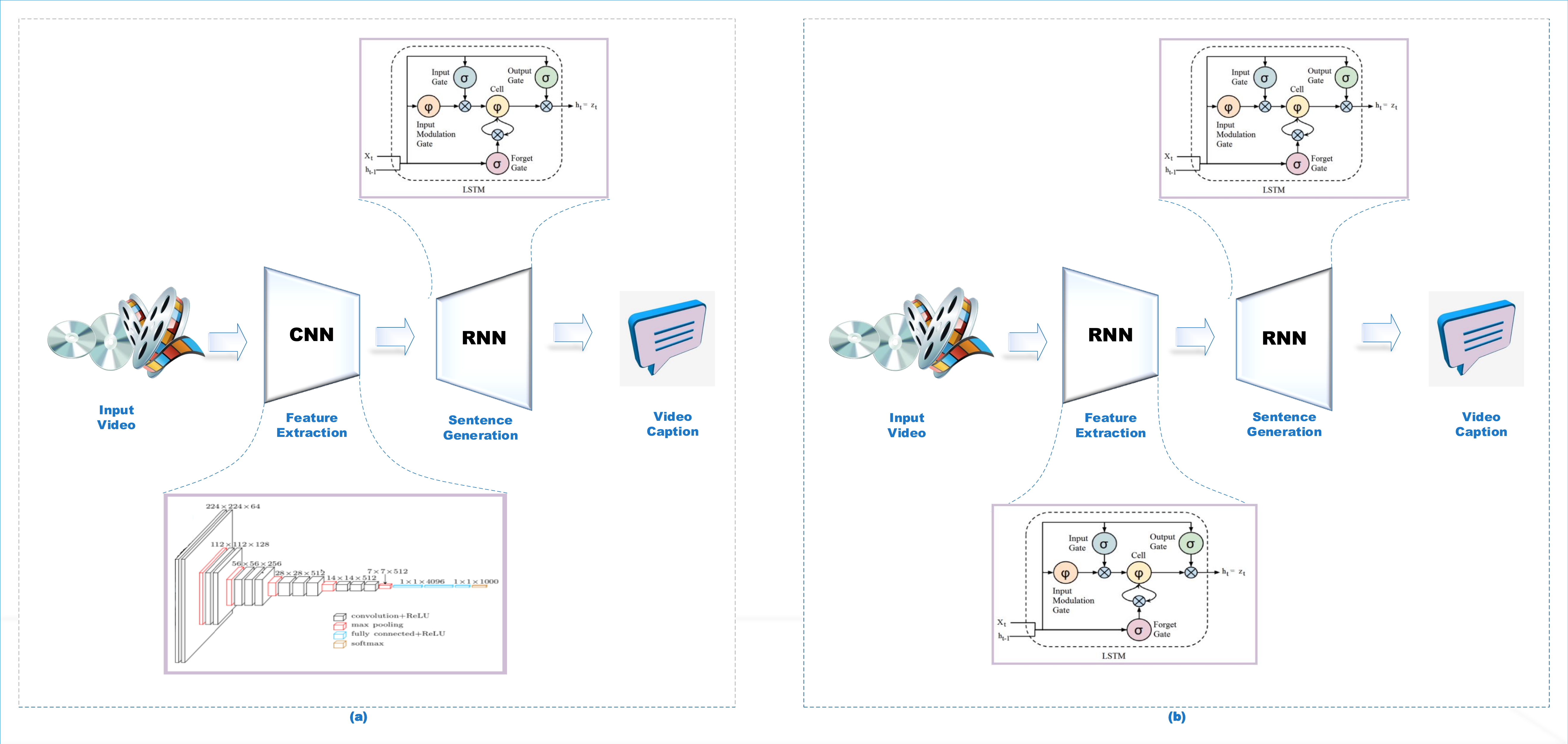}}
	\caption{General structure diagram of Video Description Model. (a) is general structure diagram of CNN-RNN architecture video description and (b) is general structure diagram of RNN-RNN architecture combination for video description.}
	\label{fig:VidDescription}
\end{figure}

\begin{figure}
	\centering
	\tmpframe{\includegraphics[width=120mm,height=45mm]{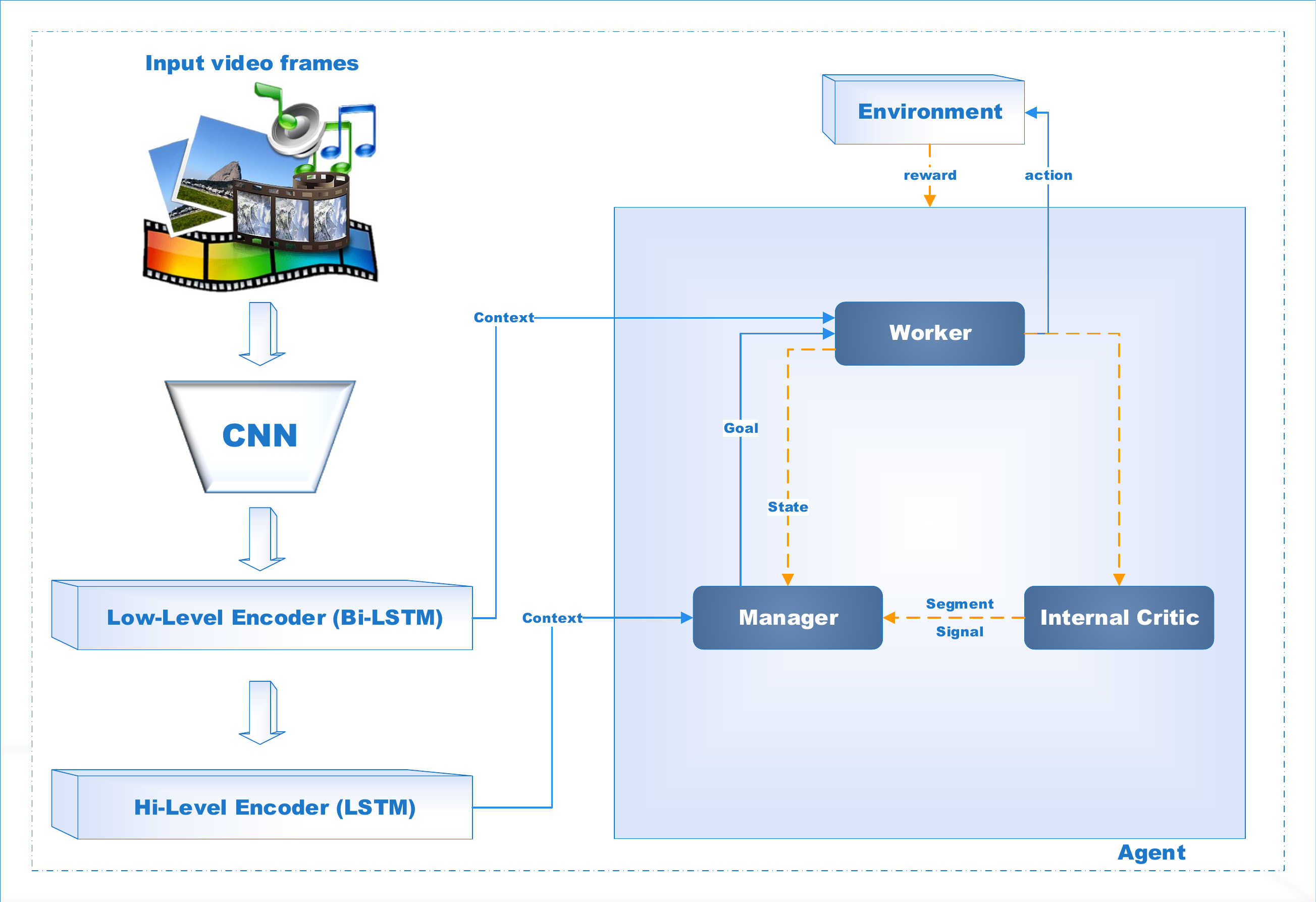}}
	\caption{General structure diagram of Video Description Deep Reinforcement Learning Architectures.}
	\label{fig:VidDescRL}
\end{figure}

Like image Description, video description is used to generate a textual description of visual contents provided through input video. It has various applications in video subtitling, visually impaired videos, video surveillance, sign language video description, and human-robot interaction. Advancements in this field open up many opportunities in various application domains. During this process, mainly two types of modalities are used, i.e., video stream and text. The general structure diagram of the video description is shown in Figures \ref{fig:VidDescription} (a) \& (b) and \ref{fig:VidDescRL}. 

During the deep learning era, many authors contribute to video description by using various methods. At the start of this DL era, some classical and statistical video description approaches are proposed based on Subject, Object, Verb (SVO) tuple methods \cite{aafaq2019video}. These SVO tuple-based methods laid the foundation for the description of the video's visual contents. Deep learning approaches can solve open domain description problems of videos. Here, DL approaches for the description of visual contents from videos are discussed in detail. There are two video description stages in deep learning techniques, i.e., visual feature extraction and text generation. In the visual feature extraction phase, various architectures are practiced like CNN, RNN, or LSTM. For the text generation phase, RNN, BRNN, LSTM, or GRU architectures are used. In this research, the video description approach is categorized based on the following architectural combinations for visual feature extraction and text generation. These approaches are comparatively analyzed according to architectures, multimedia, publication year, datasets, and evaluation metrics in Table \ref{tab:VidDescSummary}. 

\subsubsection{CNN-RNN Architectures:}
\label{CNN-RNN VidD}
Most broadly used architecture combination in the domain of video description is CNN-RNN. Figure \ref{fig:VidDescription}(a) presents the general view of the video description process by using CNN-RNN architectures. Where at the visual extraction (encoder) stage variants of CNN architectures are used and at the sentence generation (decoder) stage variants of RNN architectures are used. During deep learning era, several authors proposed techniques for describing videos that are based on this encoder, decoder combination. 
R Krishna et al. \cite{krishna2017dense} proposed a video description technique using action/event detection by applying a dense captioning mechanism. This is the first framework to detect and describe several events, but it didn't significantly improve video captioning.
B. Wang et al. \cite{wang2018reconstruction} proposed a reconstruction network for video description using an encoder-decoder reconstructor architecture, which utilizes both forward flow (from video to sentence) and backward flow (from sentence to video).  
W. Pei et al. \cite{pei2019memory} proposed an attention mechanism based encoder-decoder framework for video description. An additional memory based decoder is used to enhance the quality of video description.
N Aafaq et al. \cite{aafaq2019spatio} proposed video captioning and capitalized on Spatio-temporal dynamics of videos to extract high-level semantics using 2D and 3D CNNs hierarchically, and GRU is used for the text generation part.
S. Liu et al. \cite{liu2020sibnet} proposed SibNet; a sibling convolutional network for video description. Two architectures are used simultaneously to encode video, i.e., the content-branch to encode visual features and the semantic-branch to encode semantic features. Features of these two branches are fed together to RNN architecture by using a soft attention mechanism.

\subsubsection{RNN-RNN Architectures:}
\label{Rnn-Rnn VidD}
During the DL era, RNN-RNN is also a popular architectural combination because many authors contribute a lot by proposing various methods using this combination. Authors extract the visual content of the video by using RNN architectures instead of CNN. Figure \ref{fig:VidDescription}(b) presents the general view of the video description process by using RNN-RNN architectures. Both visual extraction (encoder) and sentence generation (decoder) stage variants of RNN architectures are used.  
H. Yu et al. \cite{yu2016video} proposed a hierarchical based two RNN architecture with attention mechanism for video description task. This approach is used for both sentence and paragraph generation for video.
M. Rahman Et al. \cite{rahman2020semantically} proposed a video captioning framework that modifies the generated context using spatial hard pull and stacked attention mechanisms. This approach illustrates that mounting an attention layer for a multi-layer encoder will result in a more semantically correct description. 
Z. Fang et al. \cite{fang2020video2commonsense} proposed a framework to generate commonsense captions of the input video. Commonsense description seeks to identify and describe the latent aspects of video like effects, attributes, and intentions. A new dataset, "Video-to-Commonsense (V2C)," is also proposed for this framework. 

\subsubsection{Deep Reinforcement Learning (DRL) architectures:}
\label{DRL VidD}
DRL is a learning mechanism where machines can learn intelligently from actions like human beings can learn from their experiences. In it, an agent is penalized or rewarded based on actions that bring the model closer to the target outcome. DRL outperforms in many ML areas and has become more popular since 2013 after the Google Deep Mind platform. The main contributions of authors using DRL architectures are;
X. Wang et al. \cite{wang2018video} proposed a hierarchical based reinforcement learning (HRL) model for describing a video. In this framework, a high-level manager module designed sub-goals and low-level worker module recognize actions to fulfill these goals. ResNet-152 extracts visual features from a video at the encoding stage, and these features are processed subsequently by low-level BLSTM \cite{schuster1997bidirectional} and high-level LSTM \cite{hochreiter1997long}. And at the decoding stage, HRL is used to generate the language description.  
Y. Chen et al. \cite{chen2018less} proposed a framework based on RL for choosing informative frames from an input video. Fewer frames are required to generate video description in this approach. To minimize the computation, on average 6$\sim$8 key/informative frames are selected to represent the complete video.  
L. Li \& B. Gong \cite{li2019end} proposed an E2E multitask RL framework for video description. In this framework, CNN and LSTM architectures are trained simultaneously for the first time. The proposed method combines RL with attribute prediction during the training process, which results in improved video description generation compared to existing baseline methods.  
J. Mun et al. \cite{mun2019streamlined} proposed a framework where an event sequence generation network is used to monitor the series of events for generated captions from the video. This framework is trained in a supervised manner, while RL further enhances the model for better context modeling.  
W. Zhang et al. \cite{zhang2019reconstruct} proposed a reconstruction network for a description of visual contents, which operates on both forward flow (from video to sentence) and backward flow (from sentence to video). Encoder-Decoder utilizes the forward flow for text description, and reconstructor (local and global) utilizes backward flow.  
W. Xu et al. \cite{xu2020deep} proposed a polishing network that utilizes the RL technique to refine the generated captions. This framework consists of word denoising and grammar checking networks for fine-tuning generated sentences. RL's policy-gradient algorithm policy optimizes the parameter of these networks.  
R. Wei et al. \cite{wei2020exploiting} proposed a framework for better exploration of RL events to generate more accurate and detailed video captions. The LSTM model is applied on video frames in a sliding window manner to leverage the local temporal information, and the temporal attention mechanism selects the most relevant sliding window to generate the most appropriate words. 

\begin{table}[] 
	\scriptsize
	\centering
	\caption{Comparative analysis of Video description models. }
	\label{tab:VidDescSummary}
	\begin{adjustbox}{width=1.1\textwidth, center=\textwidth}
		\begin{tabular}{|p{0.4cm}|p{2.6cm}|p{0.1cm}|p{3cm}|p{1.2cm}|p{3.4cm}|p{3.6cm}|}
			\hline
			\multicolumn{1}{|l|}{}&\multicolumn{1}{l|}{\textbf{Paper}}&\multicolumn{1}{l|}{\textbf{Year}}&\multicolumn{1}{l|}{\textbf{Architecture}}& \multicolumn{1}{l|}{\textbf{Multimedia}} & \multicolumn{1}{l|}{\textbf{Dataset}}&\multicolumn{1}{l|}{\textbf{Evaluation Metrics}}    \\ 
			\bottomrule
			
			\multirow{9}{*}{\cellcolor{orange!30}}  & R. Krishna et al. \cite{krishna2017dense} & 2017 & CNN/C3D, RNN/LSTM & Video, Text & ActivityNet Captions & BLEU,   CIDEr, METEOR \\
			\cellcolor{orange!30} CNN & B. Wang et al. \cite{wang2018reconstruction} & 2018 & CNN/InceptionV4, RNN/LSTM & Video, Text & MSVD,   MSR-VTT, & BLEU,   $ROUGE_{L}$, METEOR, CIDEr. \\
			\cellcolor{orange!30} to & W. Pei et al. \cite{pei2019memory} & 2019 & CNN/ResNet101/ResNeXt101, RNN/GRU & Video, Text & MSVD, MSR-VTT & BLEU,   $ROUGE_{L}$, METEOR, CIDEr. \\
			\cellcolor{orange!30} RNN & N. Aafaq et al. \cite{aafaq2019spatio} & 2019 & CNN/IRV2, 3DCNN/C3D, RNN/GRU & Video, Text & MSVD, MSR-VTT & BLEU,   ROUGE, CIDEr, METEOR \\
			\cellcolor{orange!30}& S. Liu et al. \cite{liu2020sibnet} & 2020 & CNN/GoogLeNet/Inception, RNN/LSTM & Video, Text & MSVD, MSR-VTT & BLEU,   ROUGE, CIDEr, METEOR \\
			
			\bottomrule
			\multirow{4}{*}{\cellcolor{green!15}} RNN & H. Yu et al. \cite{yu2016video} & 2016 & RNN,   RNN/GRU & Video, Text & YouTubeClips, TACoS-MultiLevel & BLEU, CIDEr, METEOR \\
			\cellcolor{green!15} to & Z. Fang et al. \cite{fang2020video2commonsense} & 2020 & CNN/ResNet-LSTM, RNN/LSTM & Video, Text & V2C,   MSR-VTT & BLEU, METEOR, $ROUGE_{L}$ \\
			\cellcolor{green!15} RNN & Md Rahman \cite{rahman2020semantically} & 2020 & CNN/VGG16-LSTM, RNN/LSTM & Video, Text & MSVD & BLEU \\
			
			\bottomrule
			\multirow{7}{*}{\cellcolor{blue!18} } 
			{\cellcolor{blue!18} }	& X. Wang et al. \cite{wang2018video} & 2018 & CNN/ResNet, RNN/BLSTM/LSTM, HRL & Video, Text & MSR-VTT, Charades & BLEU,   $ROUGE_{L}$, METEOR, CIDEr. \\
			{\cellcolor{blue!18} }	& Y. Chen et al. \cite{chen2018less} & 2018 & RNN/LSTM, RNN/GRU & Video, Text & MSVD,   MSR-VTT & BLEU,   $ROUGE_{L}$, METEOR, CIDEr. \\
			{\cellcolor{blue!18} }	& L. Li et al. \cite{li2019end} & 2019 & CNN/Inception-Resnet-V2, RNN/LSTM & Video, Text & MSVD,   MSR-VTT & BLEU,   $ROUGE_{L}$, METEOR, CIDEr. \\
			{\cellcolor{blue!18} DRL}	& J. Mun et al. \cite{mun2019streamlined} & 2019 & SST, C3D, RNN/GRU/LSTM & Video, Text & ActivityNet Captions & METEOR, BLEU, CIDEr \\
			{\cellcolor{blue!18} }	& W. Zhang et al. \cite{zhang2019reconstruct} & 2019 & CNN/InceptionV4, RNN/LSTM & Video, Text & MSVD, MSR-VTT, ActivityNet Captions & BLEU, $ROUGE_{L}$, METEOR, CIDEr. \\
			{\cellcolor{blue!18} }	& W. Xu et al. \cite{xu2020deep} & 2020 & MDP, RNN/LSTM & Video, Text & MSVD,   MSR-VTT & BLEU, $ROUGE_{L}$, METEOR, CIDEr \\
			{\cellcolor{blue!18} }	& R. Wei et al. \cite{wei2020exploiting} & 2020 & 3D CNN/ResNet, RNN/LSTM & Video, Text & MSVD,   MSR-VTT, Charades & BLEU, METEOR, CIDEr  \\
			\bottomrule
			
		\end{tabular}
	\end{adjustbox}
\end{table}

\subsection{Multimodal Visual Question Answering:} 

\begin{figure}
	\centering
	\tmpframe{\includegraphics[width=110mm,height=45mm]{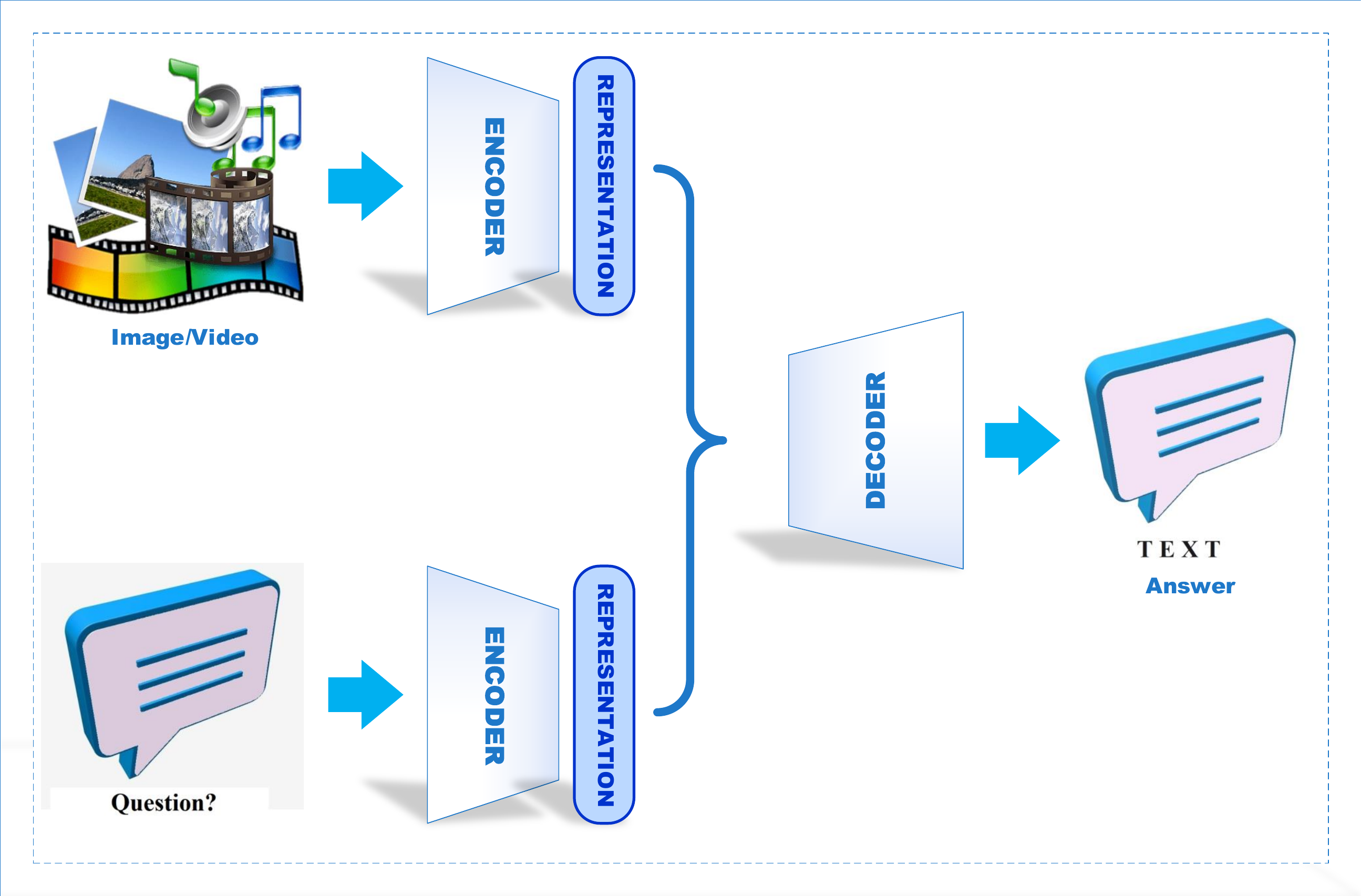}}
	\caption{General structure diagram of VQA System}
	\label{fig:VQA}
\end{figure}

VQA is an emerging technique that has piqued the interest of both the CV and NLP groups. It is a field of research about creating an AI System capable of answering natural language questions. It is a multimodal architectural system where a machine provides the answer in few words or short phrases by combining CV and NLP communities. Extracted features from input image/video and question are processed and combined to answer the question about the image as presented in Figure  \ref{fig:VQA}. CNN architectures are used to extract the image/video features while RNN architectures are used to  extract the question features. 

VQA is more complex as compared to other vision and language functions like text-to-image retrieval, video captioning, image captioning, etc. because;
\begin{itemize}
	\item[(1)] Questions asked in VQA are not specific or predetermined.
	\item[(2)] Visual information in VQA is at a high degree of dimensionality. Usually, VQA required a more thorough and detailed understanding of an image/video. 
	\item[(3)] VQA solves several CV subtasks.
\end{itemize}

Search and reasoning over image content is the significant difference in VQA. It is a multi-disciplinary AI area to process problems that belong to several sub-tasks of CV, NLP, and Knowledge Representation \& Reasoning like, object detection and recognition, attribute and scene classification, commonsense and knowledge-base reasoning, counting, activity recognition, or spatial relationships among objects.During the deep learning era, many authors contribute to the field of VQA by using various methods. These methods are grouped and presented into three groups, i.e., multimodal joint-embedding Models, multimodal attention-based models, and multimodal external knowledge-based models. Various methods of these models are comparatively analyzed in Table \ref{tab:VQASummary}.

\subsubsection{Multimodal Joint-Embedding Models (MMJEM):}
\label{VQAJoint}

MMJEM join and learn representations of multiple modalities in a common feature space. This rationale is improved further in VQA by performing more reasoning over modalities than image/video description.  
H. Ben-Younes et al. \cite{ben2017mutan} proposed a MUTAN framework for VQA. A tensor-based tucker decomposition model is used with a low-rank matrix constraint to parameterize the bi-linear relations between visual and text interpretations.  
MT Desta et al. \cite{desta2018object} proposed a framework that merges the visual features and language with abstract reasoning. High-level abstract facts extracted from an image optimize the reasoning process. 
R. Cadene et al. \cite{cadene2019murel} proposed an E2E reasoning network for VQA. This research's main contribution is introducing the MuRel cell, which produces an interaction between question and corresponding image regions by vector representation, and region relations are modeled with pairwise combinations. This MuRel network works finer than attention maps. 
B. Patro et al. \cite{patro2020robust} proposed a joint answer and textual explanation generation model. A collaborative correlated (encoder, generator, and correlated) module is used to ensure that answer and its generated explanation is correct and coherent.  
S. Lobry et al. \cite{lobry2020rsvqa} proposed a VQA framework for remote sensing data, which can be useful for land cover classification tasks. OpenStreetMap data is used for this framework, which can be further subcategorized into low and high-resolution datasets. For the remote sensing domain, these sub-datasets queries the data needed to generate the question and answers.  
Z. Fang et al. \cite{fang2020video2commonsense} proposed an open-ended VQA framework for videos using commonsense reasoning in the language part, where questions are asked about effects, intents, and attributes. 

\begin{table}[]
	\scriptsize 
	\centering
	\caption{Comparative analysis of Visual Question Answering models. }
	\label{tab:VQASummary}
	\begin{adjustbox}{width=1.1\textwidth, center=\textwidth}
	\begin{tabular}{|p{0.7cm}|p{2.6cm}|p{0.1cm}|p{4cm}|p{1.4cm}|p{3.2cm}|p{2.4cm}|}
		\hline
		\multicolumn{1}{|l|}{}&\multicolumn{1}{l|}{\textbf{Paper}}&\multicolumn{1}{l|}{\textbf{Year}}&\multicolumn{1}{l|}{\textbf{Architecture}}& \multicolumn{1}{l|}{\textbf{Multimedia}} & \multicolumn{1}{l|}{\textbf{Dataset}}&\multicolumn{1}{l|}{\textbf{Evaluation Metrics}}    \\ 
		\bottomrule
		
		\multirow{11}{*}{\cellcolor{orange!30}} & H. Ben-Younes et al. \cite{ben2017mutan} & 2017 & CNN/ResNet152, RNN/GRU & Image, Text & VQA & Accuracy \\
		\cellcolor{orange!30}  & MT. Desta et al. \cite{desta2018object} & 2018 & Faster RCNN/Resnet101, RNN/LSTM, RNN/GRU & Image, Text & CLEVR & Accuracy \\
		\cellcolor{orange!30}  \tiny \textbf{MMJEM}& R. Cadene et al. \cite{cadene2019murel} & 2019 & Faster RCNN, GRU & Image, Text & VQA v2.0, VQA-CP v2, TDIUC & Accuracy \\
		\cellcolor{orange!30}& B. Patro et al. \cite{patro2020robust} & 2020 & CNN, RNN/LSTM & Image, Text & VQA-X & BLEU, SPICE, CIDERr, METEOR, $ROUGE_{L}$ \\
		\cellcolor{orange!30}& S. Lobry et al. \cite{lobry2020rsvqa} & 2020 & CNN/ResNet152, RNN & Image, Text & OpenStreetMap (LR,HR) & Accuracy \\
		\cellcolor{orange!30}& Z. Fang et al. \cite{fang2020video2commonsense} & 2020 & CNN/ResNet-LSTM, RNN/LSTM & Video, Text & V2C-QA & TopK \\
		
		\bottomrule
		\multirow{9}{*}{\cellcolor{green!15}} &  P. Wang et al. \cite{wang2017vqa}& 2017 & CNN/VGG19-ResNet100, RNN/LSTM & Image, Text & Visual Genome QA, VQA-real & WUPS, Accuracy \\
		\cellcolor{green!15}  & Z. Yu et al. \cite{yu2017multi} & 2017 &	CNN/ResNet152, RNN/LSTM
		& Image, Text & VQA MS-COCO & Accuracy \\
		\cellcolor{green!15} \tiny \textbf{MMAM} & P. Anderson et al. \cite{anderson2018bottom} & 2018 & Faster R-CNN, RNN/GRU & Image, Text & Visual Genome, VQA v2.0 & Accuracy \\
		\cellcolor{green!15} & Z. Yu et al. \cite{yu2019deep} & 2019 & Faster R-CNN/ResNet101, RNN/LSTM & Image, Text & VQA v2.0 & Accuracy \\
		\cellcolor{green!15} & L. Li et al. \cite{li2019relation} & 2019 & Faster R-CNN/ResNet101, RNN/LSTM  & Image, Text & VQA v2.0, VQA-CP v2 & Accuracy \\
		\cellcolor{green!15} & Y. Xi et al. \cite{xi2020visual} & 2020 & CNN, RNN/LSTM  & Image, Text & DQAUAR, COCO-QA & Accuracy, WUPS \\
		\cellcolor{green!15} & W. Guo et al. \cite{guo2020re} & 2020 & Faster R-CNN, RNN/LSTM  & Image, Text & VQA v2.0 & Accuracy \\

		\bottomrule
		\multirow{7}{*}{\cellcolor{yellow!18} } &  P. Wang et al. \cite{wang2018fvqa} & 2018 & Faster R-CNN/VGG16, RNN/LSTM  & Image, Text & MS-COCO, FVQA & Accuracy TopK, WUPS \\
		{\cellcolor{yellow!18} \tiny \textbf{MMEKM}}	& M. Narasimhan et al. \cite{narasimhan2018straight} & 2018 & CNN/Faster RCNN/VGG16, RNN/LSTM & Image, Text & FVQA & Accuracy TopK \\
		{\cellcolor{yellow!18} }	& K. Marino et al. \cite{marino2019ok} & 2019 & CNN/ResNet, RNN/GRU & Image, Text & OK-VQA & Recall TopK \\
		{\cellcolor{yellow!18} }	& J. Yu et al. \cite{yu2020cross} & 2020 & Faster-RCNN, RNN/LSTM & Image, Text & FVQA, Visual7W+KB, OK-VQA & Accuracy TopK \\
		{\cellcolor{yellow!18} }	& K. Basu  et al. \cite{basu2020aqua} & 2020 & YOLO, SRE & Image, Text & CLEVR & Accuracy  \\
		\bottomrule
		
	\end{tabular}
	\end{adjustbox}
\end{table}

\subsubsection{Multimodal Attention-based Models (MMAM):}
\label{VQAAttention}
During the encoding stage, a general encoder-decoder can feed some noisy and unnecessary information at the prediction phase. MMAM are designed to improve the general baseline models to overcome this problem. The attention mechanism's main objective is to use local features of image/video and allow the system to assign priorities to the extracted features from different regions. In an image, the attention portion identified salient regions, and then the language generation part is more focused on those regions for further processing. This concept is also used in VQA to enhance model's performance by focusing on a specific part of an image according to the asked question. The attention step does some additional processing in the reasoning part by focusing on a particular part of the image/video before further language computations.

During the DL era, various approaches are proposed for VQA tasks based on attention mechanisms, 
like P. Wang et al. \cite{wang2017vqa} proposed a VQA framework based on a co-attention mechanism. In this framework, the co-attention mechanism can process facts, images, and questions with higher order. 
Z. Yu et al. \cite{yu2017multi} proposed a factorized bi-linear pooling method with co-attention learning for the VQA task. Bilinear pooling methods outperform the conventional linear approaches, but the practical applicability is limited due to their high computational complexities and high dimensional representations.
P. Anderson et al. \cite{anderson2018bottom} proposed bottom-up and top-down attention based framework for VQA. This attention mechanism enables the model to calculate feature based on objects or salient image regions. 
Z. Yu et al. \cite{yu2019deep} proposed a deep modular co-attention network for the VQA task. Each modular co-attention layer consists of question guided self-attention mechanism of images using the modular composition of both image and question attention units. 
L. Li et al. \cite{li2019relation} proposed a relation aware graph attention mechanism for VQA. This framework encodes the extracted visual features of an image into a graph, and to learn question-adaptive relationship representations, a graph attention mechanism modeled inter-object relations.  
W. Guo et al. \cite{guo2020re} proposed a re-attention based mechanism for VQA. The attention module correlates the pair of object-words and produces the attention maps for question and image with each other's guidance.

\subsubsection{Multimodal External Knowledge Bases Models (MMEKM):}
\label{VQAknowledge}
Traditional multimodal joint embedding and attention-based models only learn from the information that is present in training sets. Existing datasets do not cover all events/activities of the real world. Therefore, MMEKM is vital to coping with real-world scenarios. Performance of VQA task is more increasing by linking knowledge bases (KB) databases to VQA task.  
Freebase \cite{bollacker2008freebase}, DBPedia \cite{auer2007dbpedia}, WordNet \cite{miller1995wordnet}, ConceptNet \cite{liu2004conceptnet}, and WebChild \cite{tandon2014webchild} are extensively used KB. A robust VQA framework requires access to broad information content from KB. It has been effectively integrated into the VQA task by embedding the various entities and relations. 

During the DL era, various external KB methods are proposed for VQA tasks. 
P. Wang et al. \cite{wang2018fvqa} proposed another framework for the VQA task names "Fact-based VQA (FVQA)" that uses data-driven approaches and LSTM architecture to map image/question queries. FVQA framework used DBPedia, ConceptNet, and WebChild KB. 
M. Narasimhan \& AG. Schwing \cite{narasimhan2018straight} proposed a framework for the VQA task using external knowledge resources that contain a set of facts. This framework can answer both fact-based and visual-based questions. 
K. Marino et al. \cite{marino2019ok} proposed an outside knowledge dataset for VQA, which contains more than 14,000 questions. This dataset contains several categories like sports, science and technology, history, etc. This dataset requires external resources to answer, instead of only understanding the question and image features.  
K. Basu et al. \cite{basu2020aqua} proposed a commonsense based VQA framework. In this framework, the image's visual contents are extracted and understood by the YOLO framework and represented in the answer set program.  Semantic relations features and additional commonsense knowledge answer the complex questions for natural language reasoning.
J. Yu et al. \cite{yu2020cross} proposed a framework in which visual contents of an image is extracted and processed in multiple perspectives of knowledge graph like semantic, visual, and factual perspectives. 

\begin{figure}
	\centering
	\tmpframe{\includegraphics[width=120mm,height=45	mm]{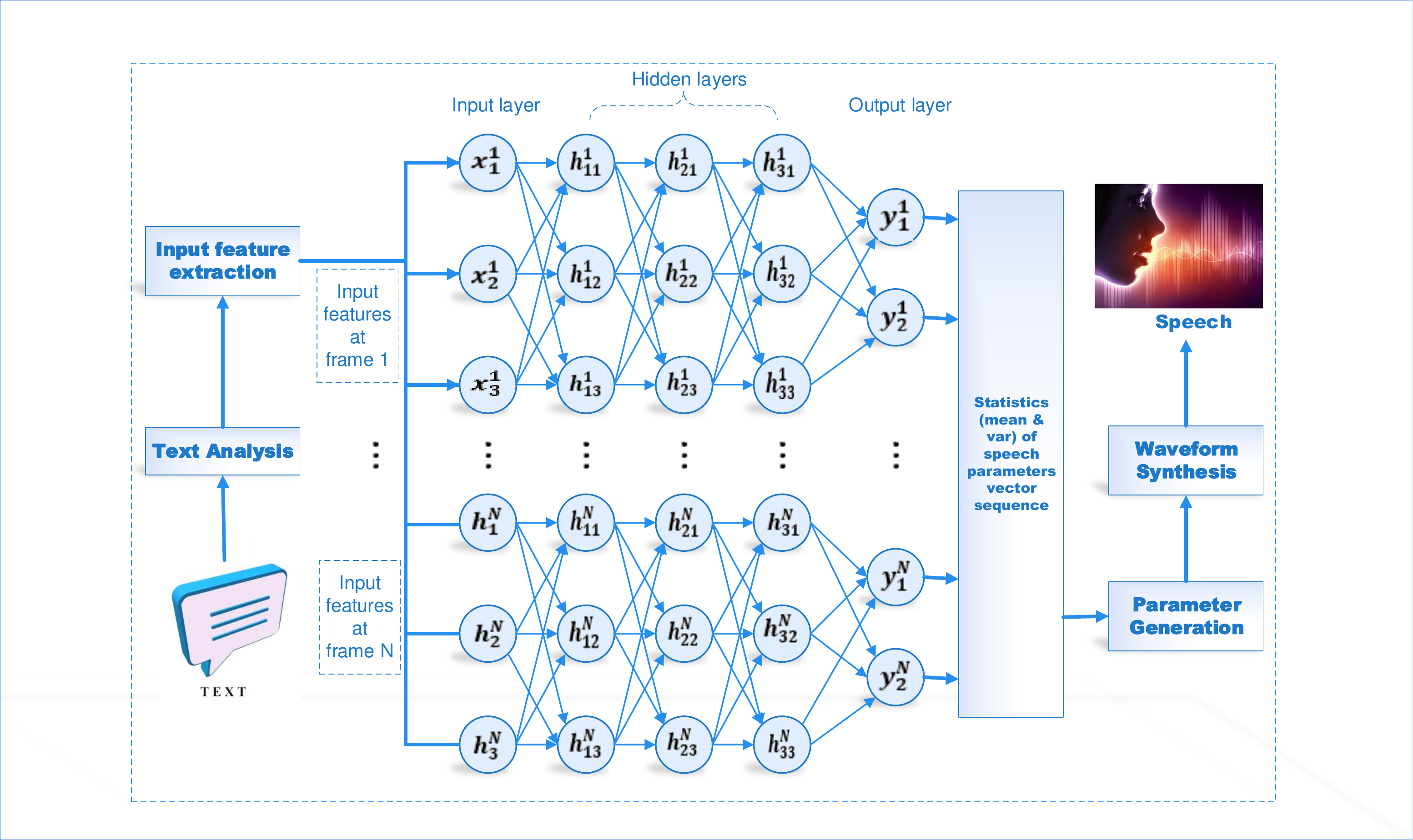}}
	\caption{General structure diagram of Deep Learning Text To Speech framework using DNN architecture.}
	\label{fig:SpeechSynthesis}
\end{figure}

\subsection{Multimodal Speech synthesis:}
The most important aspect of human behavior is communication (write/speak). Humans can communicate using natural language by text and speech, representing the written and vocalized form of natural language, respectively. The latest research in language and speech processing helps systems talk like a human being. Speech synthesis is the complicated process of generating natural language spoken by a machine. Natural language text modality is converted into its respective spoken waveform modality in real-time by the Text To Speech (TTS) system. Various applications are introduced in the real world using speech synthesis, like human-computer interactive systems, screen readers, telecommunication and multimedia applications, talking toys games, etc. The main research objective of TTS systems these days is to produce a sound like a human's. Therefore, various aspects are used for the evaluation of the TTS system's quality, such as naturalness (quality with the perspective of generated speech timing structure, rendering emotions and pronunciation), intelligibility (in a sentence, quality of each word being produced), synthetic speech preferences (choice of a listener in term of voice and signal quality for better TTS system) and human perception factors like comprehensibility (understanding quality of received messages). Articulatory TTS, Concatenative TTS, Formant TTS, Parametric TTS, and Deep Learning TTS are the main categories of the speech synthesis process. This section discusses recent research trends and advancements of Deep Learning TTS.

\subsubsection{Deep Learning TTS (DLTTS):}
\label{DLTTS}
In DLTTS frameworks, DNN architectures model the relationship between text and their acoustic realizations. The main advantage of DLTTS is the development of its extensive features without human prepossessing. Also, the naturalness and intelligibility of speech are improved using these systems. Text to speech synthesis process is explained in the general structure diagram of Deep Learning Text To Speech frameworks using DNN architectures, shown in the Figure \ref{fig:SpeechSynthesis}. Comparative analysis of these approaches is shown in Table \ref{tab:TTS}.

Y. Wang et al. \cite{wang2017tacotron} proposed "Tacotron," a sequence 2 sequence TTS framework that synthesizes speech from text and audio pairs. Encoder embeds the text that extracts its sequential representations. The attention-based decoder process these representations, and after that, post-processing architecture generates the synthesized waveforms.
In an another research, "Deep Voice" model using DNN architecture are proposed by SO Arik et al. \cite{arik2017deep} synthesizes audio from characters. This model consists of five significant blocks for the production of synthesized speech from text. The computational speed is increased compared to existing baseline models because the model can train without human involvement.
A. Gibiansky et al. \cite{gibiansky2017deep} proposed a Deep Voice-2 architecture. This framework is designed to improve existing state-of-the-art methods, i.e., Tacotron and Deep Voice-1, by extending multi-speaker TTS through low-dimension trainable speaker embedding.
In third version of Deep Voice, W. Ping et al. \cite{ping2017deep} proposed a neural TTS system based on the fully convolutional model with an attention mechanism. This model performs parallel computations by adapting Griffin-Lim spectrogram inversion, WORLD, and WaveNet vocoder speech synthesis.  
"Parallel WaveNet" an advanced version of WaveNet proposed by A. Oord et al. \cite{oord2018parallel} using the probability density distribution method to train networks. A teacher and a student WaveNet are used parallelly in this model. 
SO Arik et al. \cite{arik2018neural} proposed a neural voice cloning system that learns human voice from fewer samples. For that purpose, two techniques, i.e., speaker adaptation and encoding, are used together.
Y. Taigman et al. \cite{taigman2018VoiceLoopVF} proposed a VoiceLoop framework for a TTS system. This model can deal with un-constrained voice samples without the need for linguistic characteristics or aligned phonemes. This framework transformed the text into speech from voices using a short-shifting memory buffer.  
J. Shen et al. \cite{shen2018natural} proposed "Tacotron2. It is neural TTS architecture used to synthesize speech directly from the text. A recurrent based sequence to sequence feature prediction network can map characters to spectrogram, and then these spectrogram are used to synthesize waveforms by using a modified version of WaveNet vocoder.  
Parallel Tacotron is another brilliant invention in recent times for neural TTS approach proposed by I. Elias et al. \cite{elias2020parallel}. During inference and training processes, this approach is highly parallelizable to achieve optimum synthesis on modern hardware. One to many mapping nature of VAE enhance the performance of TTS and also improves its naturalness.

\begin{table}
	\scriptsize
	\centering
	\caption{Comparatively analysis of speech synthesis (DLTTS) models. }
	\label{tab:TTS}
	\begin{adjustbox}{width=1.1\textwidth, center=\textwidth}
		\begin{tabular}{|p{2.3cm}|p{0.4cm}|p{2.8cm}|p{1.2cm}|p{4.2cm}|p{2.1cm}|}
			\bottomrule
			\cellcolor{blue!20}\textbf{Paper}  & \cellcolor{blue!20}\textbf{Year} &  \cellcolor{blue!20}\textbf{Architecture}  & \cellcolor{blue!20}\textbf{Multimedia} & \cellcolor{blue!20}\textbf{Dataset} & \cellcolor{blue!20}\textbf{Evaluation Metrics} \\
			\bottomrule
			Y. Wang et al. \cite{wang2017tacotron} & 2017 & Bidirectional-GRU/RNN & Text, Audio & North American English & MOS \\
			SO Arik et al. \cite{arik2017deep} & 2017 & GRU/RNN & Text, Audio & Internal English speech database, Blizzard & MOS \\
			A. Gibiansky et al. \cite{gibiansky2017deep} & 2017 & GRU/RNN & Text, Audio & Internal English speech database & MOS \\
			W. Ping et al. \cite{ping2017deep} & 2017 & Griffin-Lim, WORLD, WaveNet vocoder & Text, Audio & Internal English speech database, VCTK, LibriSpeech & MOS \\
			A. Oord et al. \cite{oord2018parallel} & 2018 & CNN & Audio & North American English, Japanese & MOS \\
			SO Arik et al. \cite{arik2018neural} & 2018 & Griffin-Lim vocoder & Text, Audio & LibriSpeech, VCTK & Accuracy, EER, MOS \\
			Y. Taigman et al. \cite{taigman2018VoiceLoopVF} & 2018 & GMM & Text, Audio & Lj speech, Blizzard, VCTK & MOS, MCD, TopK \\
			J. Shen et al. \cite{shen2018natural} & 2018 & STFT, BiLSTM & Text, Audio & Internal US English & MOS \\
			I. Elias et al. \cite{elias2020parallel} & 2020 & GLU, VAE, LSTM & Text, Audio & Proprietary speech & MOS \\
			\bottomrule
		\end{tabular}
	\end{adjustbox}
\end{table}

\subsection{Other MMDL applications:}
\label{others apps}

\subsubsection{Multimodal Emotion Recognition (MMER):}
\label{EmotionRecog}

\begin{figure}
	\centering
	\tmpframe{\includegraphics[width= 160mm, height=45mm]{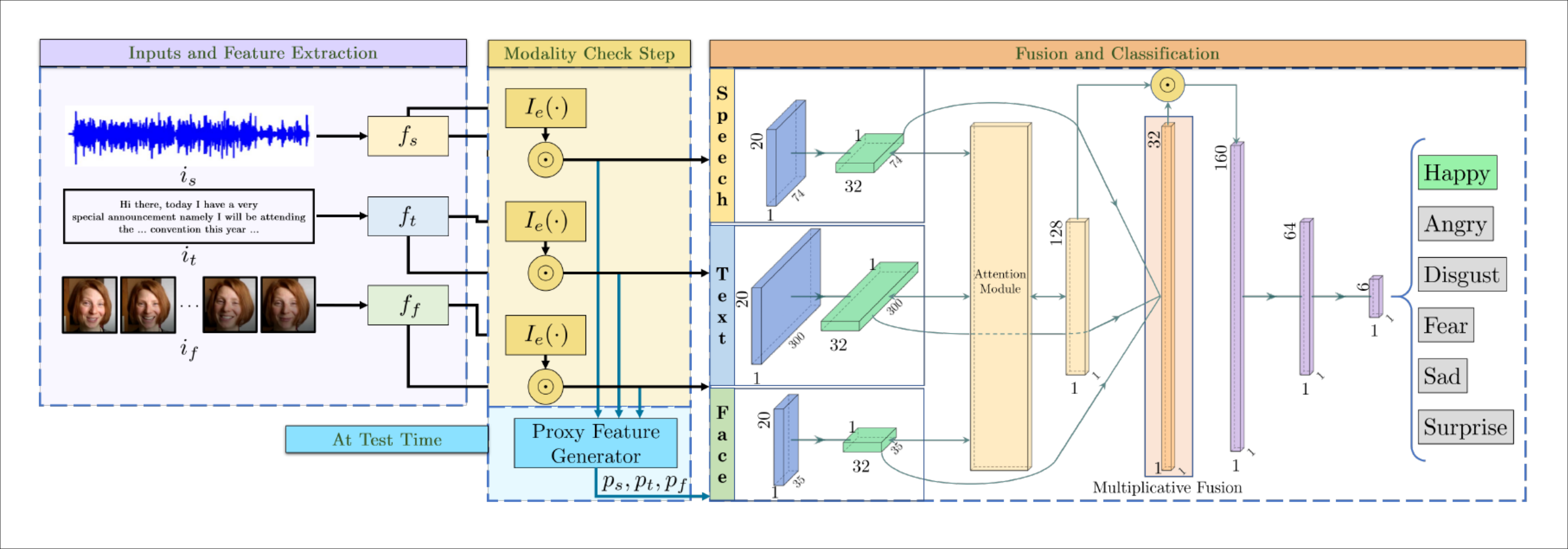}}
	\caption{General structure diagram of Multimodal Emotion Recognition using Facial, Textual, and Speech Cues \cite{mittal2020m3er}.}
	\label{fig:EmotionDetection}
\end{figure}

Emotions of human is one of the approach for expressing feelings. MMER is enormously vital for enhancing the interaction experience between humans and computers. The task of ML is to empower computers to learn and identify new inputs from training datasets, hence it can be used efficiently to detect, process, respond, understand and recognize emotions of human through computers training. Thus, the primary purpose of affective computing is to provide machines/systems emotional intelligence ability. It has several research areas such as learning, health, education, communication, gaming, personalized user interface, virtual reality and information retrieval. Multimodal emotion recognition framework can be developed on an AI/ML based prototypes designed to extract and process emotion information from various modalities like speech, text, image, video, facial expression, body gesture, body posture, and physiological signals. The general structure diagram of multiplicative multimodal emotion recognition using facial, textual, and speech cues are shown in Figure \ref{fig:EmotionDetection}.

During the DL era, various authors contribute to emotion recognition by using different architecture and multiple modalities. Like,  
Y. Huang et al. \cite{huang2017fusion} proposed a fusion method for emotion recognition using two modalities, i.e., facial expression and electroencephalogram (EEG) signals. A NN classifier detects happiness, neutral, sadness, and fear states of emotions.  
D. Nguyen et al. \cite{nguyen2017deep} proposed another approach for emotion recognition by using audio and video streams. In this method, a combination of C3D and DBN architectures is used to model Spatio-temporal information and representation of video audio streams.
S. Tripathi et al. \cite{tripathi2018multi} proposed a multimodal emotion recognition framework using various modalities like, text, speech, face expression, and hands movement on the IEMOCAP dataset. The fusion of modalities is performed only at the final layer to improve emotion recognition performance.
D. Hazarika et al. \cite{hazarika2018icon} proposed an emotion recognition framework from video conversations. This method can generate self and interpersonal affective summaries from conversations by using contextual information extracted from videos. They also proposed another framework for detecting emotions using the attention mechanism \cite{hazarika2018self}. In this framework, audio and textual modalities are used for the detection of emotions.  
M. Jaiswal et al. \cite{jaiswal2019controlling} analyzes the change of emotional expressions under various stress levels of an individual. The performance of this task is affected by the degree of lexical or acoustic features.  
L. Chong et al. \cite{chong2019emochat} proposed a new online-chatting system called "EmoChat" to automatically recognize the user's emotions and attach identified emotion automatically and the sent message within a short period. During the chatting user can know each other's emotions in this network.
M. Li et al. \cite{li2020multistep} proposed a multi-step deep system for reliable detection of emotions by using collected data that contains invalid data as well. Neural networks are used to filter this invalid data from videos and physiological signals by using continuity and semantic compatibility.  
H. Lai et al. \cite{lai2020different} proposed a model for emotion recognition in interactive conversations. In this model, different RNN architectures with variable contextual window sizes differentiate various aspects of contexts in conversations to improve the accuracy. 
RH Huan et al. \cite{huan2020video} proposed a model using attention mechanism. In this method, for better use of visual, textual and audio features for video emotion detection, bidirectional GRU is cascaded with an attention mechanism. The attention function is used to work with various contextual states of multiple modalities in real-time.  
Y Cimtay et al. \cite{cimtay2020cross} proposed a hybrid fusion method for emotion recognition using three modalities, i.e., facial expression, galvanic skin response (GSR) and EEG signals. This model can identify the actual emotion state when it is prominent or concealed because of natural deceptive face behavior. 

\begin{table}[]
	\scriptsize 
	\centering
	\caption{Comparative analysis of MMER and MMED models. }
	\label{tab:OthersSummary}
	\begin{adjustbox}{width=1.1\textwidth, center=\textwidth}
	\begin{tabular}{|p{0.6cm}|p{2.2cm}|p{0.1cm}|p{3.2cm}|p{2.4cm}|p{2.8cm}|p{3cm}|}
		\hline
		\multicolumn{1}{|l}{}&\multicolumn{1}{l}{\textbf{Paper}}&\multicolumn{1}{|l}{\textbf{Year}}&\multicolumn{1}{|l}{\textbf{Architecture}}& \multicolumn{1}{|l}{\textbf{Multimedia}} & \multicolumn{1}{|l}{\textbf{Dataset}}&\multicolumn{1}{|l|}{\textbf{Evaluation Metrics}}    \\ 
		\bottomrule
		
		\multirow{12}{*}{\cellcolor{red!20}} & Y. Huang et al. \cite{huang2017fusion} & 2017 & SVM & Physiological signals, facial expression & Manual offline data collection & Weighted Accuracy \\
		\cellcolor{red!20}& D. Nguyen et al. \cite{nguyen2017deep} & 2017 & C3D, DBN & Video, Audio & eNTERFACE Audio-visual & Weighted Accuracy \\
		\cellcolor{red!20}& D. Nguyen et al. \cite{nguyen2018deep} & 2018 & C3D, MCB, DBN & Video, Audio&eNTERFACE Audio-visual, FABO &	Weighted Accuracy \\
		\cellcolor{red!20} & S. Tripathi et al. \cite{tripathi2018multi}&2018&MFCC, RNN/LSTM&Audio, Text, Facial expressions, Body gesture	&IEMOCAP&	Weighted Accuracy\\
		\cellcolor{red!20}  & D. Hazarika et al. \cite{hazarika2018icon}&2018&CNN, 3D-CNN, RNN/GRU&Text, Audio, Video&	IEMOCAP, SEMAINE&	Weighted Accuracy, F1-Score, MAE \\
		\cellcolor{red!20} & 	D. Hazarika et al. \cite{hazarika2018self}&2018&CNN, OpenSMILE, MFCC&Text, Audio&	IEMOCAP &	Weighted Accuracy, F1-Score, Recall\\
		\cellcolor{red!20} MMER & M. Jaiswal et al. \cite{jaiswal2019controlling}& 2019 & GRU & Physiological signals, Text & MuSE, IEMOCAP, MSP-Improv & Weighted Accuracy\\
		\cellcolor{red!20}  & L. Chong et al. \cite{chong2019emochat}&2019&TextCNN, Mini-Xception, HMM&Text, Facial expression 	&FER2013&	Weighted Accuracy\\
		\cellcolor{red!20}& M. Li et al. \cite{li2020multistep}&2020&C3D, DBN, SVM-RBF&Video, Physiological signals&	RECOLA&	Weighted Accuracy, F1-Score, Recall\\
		\cellcolor{red!20}& H. Lai et al. \cite{lai2020different}&2020&CNN, 3D-CNN, RNN/GRU&Text, Audio, Video&	IEMOCAP, AVEC	&Accuracy, F1-Score\\
		\cellcolor{red!20}& RH Huan et al. \cite{huan2020video}&2020&RNN/GRU&Video, Text, Audio&	CMU-MOSI, POM& 	MAE, Accuracy, F1-Score, Precision, Recall\\
		\cellcolor{red!20}& Y Cimtay et al. \cite{cimtay2020cross}&2020&CNN/InceptionResnetV2& Facial expressions, Physiological signals&	LUMED-2, DEAP&	Weighted Accuracy\\
		\bottomrule
		
		\multirow{5}{*}{\cellcolor{purple!15}} & Y. Gwon et al. \cite{gwon2016multimodal} & 2016 & MFCC, CNN & Audio, Video & TRECVID MED & mAP, Accuracy \\
		\cellcolor{purple!15}  & Y. Gao et al. \cite{gao2017event} & 2017 & CNN, HMM & Image, Text & Brand-Social-Net & Recall, precision, F1-Score  \\
		\cellcolor{purple!15} MMED & S. Huang et al. \cite{huang2018learning} & 2018 & CRBM, CDBN, SVM & Video & UCSD, Avenue & AUC, EER, EDR, Accuracy \\
		\cellcolor{purple!15}  & P. Koutras et al. \cite{koutras2018exploring} & 2018 &	CNN, C3D & Audio, Video & COGNIMUSE  & AUC \\
		\cellcolor{purple!15}  & Z. Yang et al. \cite{yang2019shared} & 2019 & CNN/VGGNet-AlexNet & Image, Text & MED, SED & NMI, F1-Score \\
		\bottomrule
		
	\end{tabular}
\end{adjustbox}
\end{table}

In this section, recent existing methods for emotion recognition and analysis using multimodal DL are presented. Comparative analysis of these techniques is shown in Table \ref{tab:OthersSummary}. This literature review on emotion detection is based on text, audio, video, physiological signals, facial expressions and body gestures modalities. It is clearly shown from an empirical study that automatic emotion analysis is feasible and can be very beneficial in increasing the accuracy of the system response and enabling the subject's emotional state to be anticipated more rapidly. Cognitive assessment and their physical response are also analyzed along with primary emotional states.

\subsubsection{Multimodal Event Detection:}
\label{EventDetection}

Due to the popularity of media sharing on the internet, users can easily share their events, activities, and ideas anytime. The aim of multimodal event detection (MMED) systems is to find actions and events from multiple modalities like images, videos, audio, text, etc. According to statistics, Million of tweets are posted per day, and similarly, YouTube users post more than 30,000 hours of videos per hour. Hence, in many CV applications, automatic event and action detection mechanisms are required from this large volume of user-generated videos. Finding events and actions from this extensive collection of data is a complex and challenging task. It has various applications in the real-world like disease surveillance, governance, commerce, etc and also helps internet users to understand and captures happenings around the world. Summary of various MMED methods are presented in Table \ref{tab:OthersSummary}. 

Researchers have contributed a lot during the DL era and proposed many methods for event and action detection using multiple modalities. Like,  
Y. Gwon et al. \cite{gwon2016multimodal} proposed a multimodal sparse coding scheme for the learning of multiple modalities features representation. In this approach, these feature representations are used to detect events from audio and video media.   
Y. Gao et al. \cite{gao2017event} proposed a method for event classification via social tracking and deep learning in microblogs. Images and text media are fed to a multi-instance deep network to classify events in this framework.  
S. Huang et al. \cite{huang2018learning} proposed an unsupervised method for the detection of anomaly events from crowded scenes using DL architectures. In this model, visual, motion map and energy features are extracted from video frames. A multimodal fusion network utilized these features for the detection of anomaly events by using the SVM model.
In another research to detect salient events from videos, P. Koutras et al. \cite{koutras2018exploring} employed CNN architecture using audio and video modalities. In this framework, a CNN architecture based on C3D nets is used to detect events from a visual stream, and a 2D-CNN architecture is used to detect events from the audio stream. Experimental results show the improvement of the proposed method performance over the baseline methods for salient event detection.
Z. Yang et al. \cite{yang2019shared} proposed a framework for event detection from multiple data domains like social and news media to detect real-world events. A unified multi-view data representation is built for image and text modalities from social and news domains in this framework. Class wise residual units are formulated to identify the multimedia events.

\section{MMDL Architectures:}
\label{architectMMDL}

Multimodal processing, learning and sense have a range of deep learning models. Deep learning is a hierarchical computation model, learns the multilevel abstract representation of the data based on the Artificial Neural Networks (ANN). Hierarchical representation of DL can learn automatically for each modality rather than manually designing particular features of modalities. In this section, a brief explanation of different architectures is presented, which is used in various applications mentioned in section \ref{AppsMMDL}. These architectures are categorized into probabilistic graphical models, ANN and miscellaneous architectures groups. 

\begin{figure}
	\centering
	\tmpframe{\includegraphics[width=140mm, height=45mm]{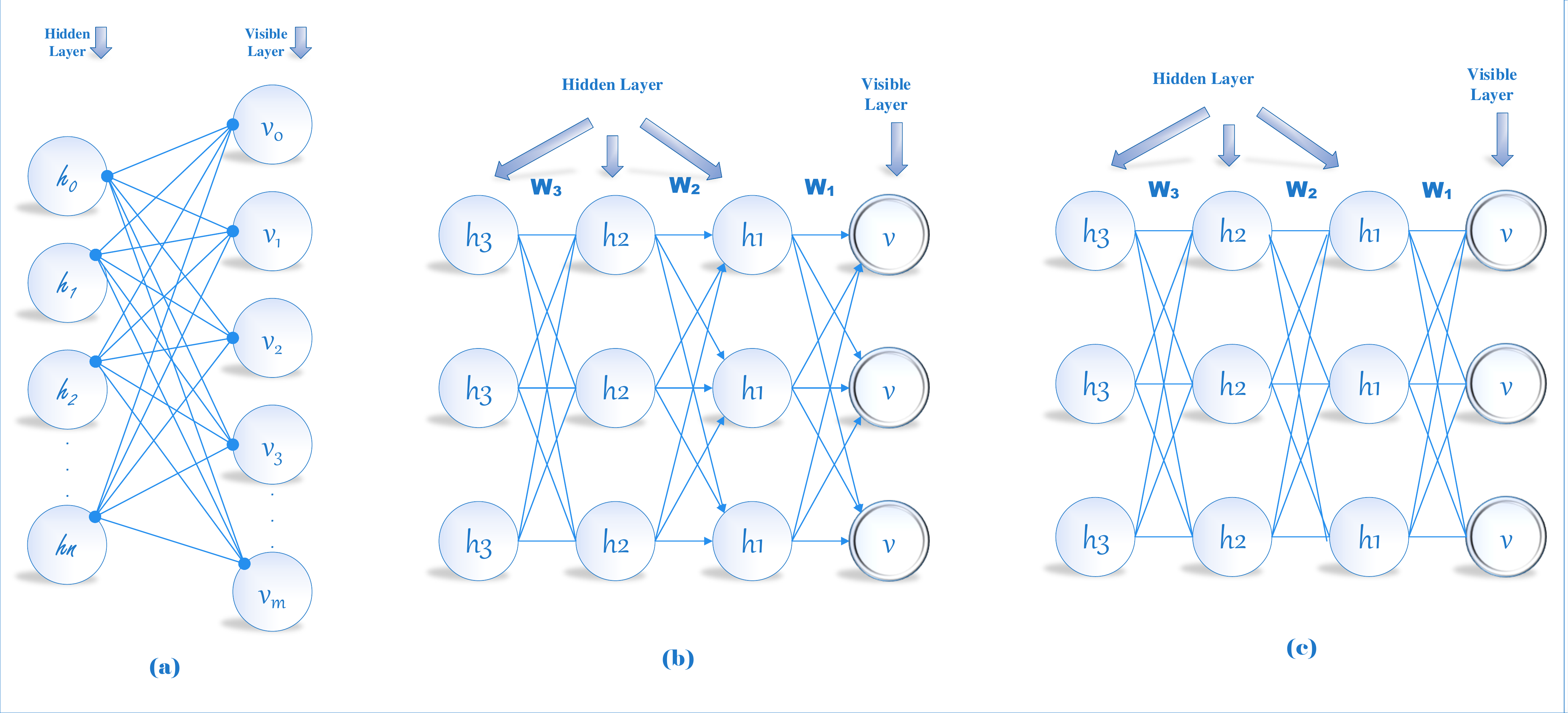}}
	\caption{Structure diagrams of Probabilistic Graphical models: (a) Restricted Boltzmann Machine (RBM), (b) Deep Belief Network (DBN); Partially directed connections between layers and (c) Deep Boltzmann Machine (DBM); Fully undirected connections between layers}
	\label{fig:RbmDbnDbm}
\end{figure}

\subsection{Probabilistic graphical models:}

In MMDL architectures, probabilistic graphical models includes Restricted Boltzmann machine (RBM) \cite{smolensky1986information}, Deep Belief Networks (DBN) \cite{hinton2006fast}, Deep Boltzmann Machines (DBM) \cite{salakhutdinov2009deep}, and Variational Auto-Encoders \cite{hou2017deep}.
\textbf{Restricted Boltzmann Machines (RBM)}
The most popular sub-class of the Boltzmann Machine \cite{ackley1985boltzmann} is RBM. RBM is a stochastic and generative model with two layers; one is the stochastic visible layer $v \in \{0, 1\}^D$, and the other is a stochastic hidden layer $h \in \{0, 1\}^F$ that can learn the distribution of training data. It is an undirected graphical modal whose connections are restricted so that it must form a bipartite graph by its neurons. The structure diagram of RBM is shown in Figure \ref{fig:RbmDbnDbm}(a). RBM can be broadly extended to its respective variants when considering visible sparse count or real-valued data like, replicated softmax RBM \cite{pang2015mutlimodal}, Fuzzy RBM (FRBM) \cite{chen2015fuzzy}, Fuzzy Removing Redundancy RBM (F3RBM) \cite{lu2019fuzzy}, Gaussian-Bernoulli RBM (GBRBM) \cite{tan2019parallel}, and Normalized RBM (NRBM) \cite{aamir2019efficient}.
\textbf{Deep Belief Networks (DBN)} is another probabilistic graphic model that offers a joint probability distribution over the observable labels and data \cite{hinton2006fast}. It is a graphical generative model that learns how to extract a deep hierarchical representation of the training data. DBN is composed by stacking multiple RBMs and train them in an unsupervised greedy manner by using a contrastive divergence procedure. Initially, DBN employs an efficient layer-by-layer greedy learning strategy to initialize deep networks and tunes all weights with the desired output in parallel. At the top layers, it has undirected connections, and at the lower layers, it has directed connections. The structure diagram of DBN is shown in Figure \ref{fig:RbmDbnDbm}(b). DBN has been widely used in various fields to address representation learning, semantic hashing and data dimension reduction. DBN can be broadly extended to its respective variants like Fuzzy Deep Belief Network (FDBN) \cite{zhou2014fuzzy}, Discriminative Deep Belief Networks (DDBN) \cite{ma2017discriminative}, Competitive Deep Belief Networks (CDBN) \cite{yang2018competitive}, and Adaptive Fractional Deep Belief Network (AFDBN) \cite{mannepalli2017novel}.

\textbf{Deep Boltzmann Machines (DBM)} is another probabilistic graphic model; RBM is used as the building block for DBM model. The main difference of DBM model with DBN is that connections between all layers are undirected. It is a completely generative model that can extract features from data with some missing modalities. The right selection of interactions between hidden and visible layers can result in a more tractable version of DBM. It can capture several layers of complex input data representations and are suitable for unsupervised learning because they can be trained on unlabeled data. Nevertheless, it can also be tuned in a supervised manner for a specific task. In DBM, all the layers' parameters can be optimized jointly by following the approximate gradient of lower bound variation. This is especially helpful in cases where heterogeneous data learning models from different modalities.The structure diagram of DBM is shown in Figure \ref{fig:RbmDbnDbm}(c), it can be broadly extended to its respective variants like robust spike-and-slab deep Boltzmann machine (RossDBM) \cite{zhang2020robust}, Batch Normalized Deep Boltzmann Machines (BNDBM) \cite{vu2018batch}, Contractive Slab and Spike Convolutional Deep Boltzmann Machine (CssCDBM) \cite{xiaojun2018contractive}, and Mean Supervised Deep Boltzmann Machine (msDBM) \cite{nagpal2019expression}.
And \textbf{Variational Auto-Encoders (VAE)} \cite{hou2017deep} are directed probabilistic graphical models, a DL approach for latent representations learning. VAE is one of the most famous techniques for complex distributions in unsupervised learning approaches. It provides a probabilistic way of defining latent space observations and ensures that latent space contains good properties to enable the generative process. VAE is also considered a high-resolution network because it is used to design complex models with large datasets. It shows promising progress in producing complex data like faces, handwritten digits, segmentation, the future prediction from static images and speech synthesis, etc. 

\subsection{Artificial Neural Network (ANN):}
 
 \begin{figure}
 	\centering
 	\tmpframe{\includegraphics[width= 160mm, height=35mm]{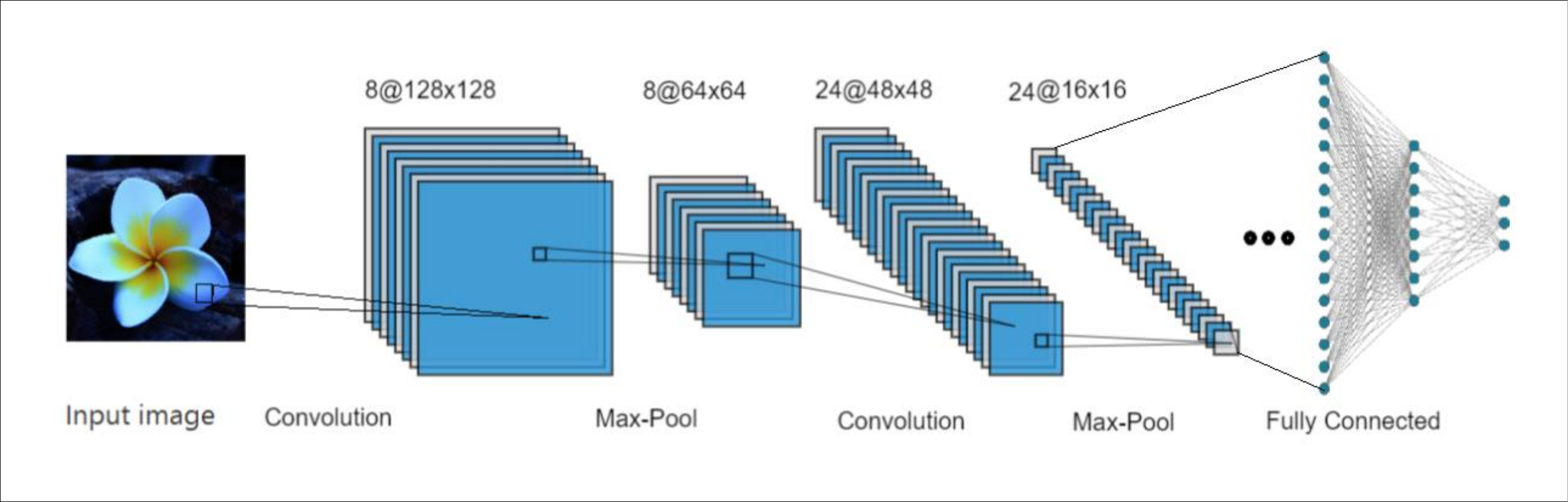}}
 	\caption{General Diagram of Convolutional Neural Network architecture.}
 	\label{fig:CNN}
 \end{figure}

\begin{figure}
	\centering
	\tmpframe{\includegraphics[width= 160mm, height=35mm]{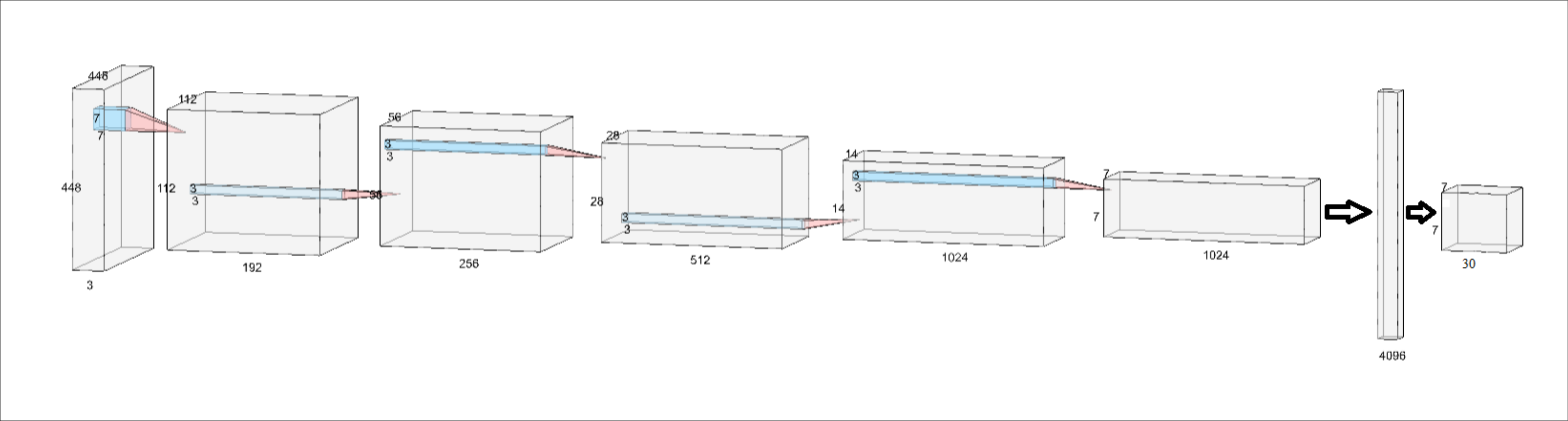}}
	\caption{General Diagram of You Only Look Once architecture.}
	\label{fig:YOLO}
\end{figure} 

A biological neuron's computational powers inspire an ANN. These NNs are the origin of AI and solve complex problems by efficiently utilizing its self-learning capabilities. It is used in various activities, including CV, speech recognition, social network filtering, medical diagnosis, machine translation, video games, etc. According to the utilization of ANN architectures in various applications discussed in section \ref{AppsMMDL}, three basic ANN categories have been briefly explained in this article, i.e., CNN, RNN, and YOLO architectures. \textbf{Convolutional Neural Network (CNN)} architecture is a DL algorithm composed of input, hidden, and output layers to solve complex patterns. These hidden layers consist of convolutional, pooling and fully connected layers.  Convolutional layers contain distinct filters, which depict input data in fewer dimensional slices to generate feature maps. Then pooling layer performs the sub-sampling on these feature maps by reducing their dimensionality. Pre-processing required for CNN architecture is much less in contrast to other classification algorithms. CNN was introduced in 1989 when LeCuN's work on the processing of grid-like time series and image data \cite{lecun1989backpropagation}. CNN is used in various application areas like image segmentation, image classification, video processing, object detection, Speech recognition, NLP, recommender system, anomaly detection, medical image analysis, drug discovery, media recreation etc. A general Diagram of Convolutional Neural Network architecture is shown in Figure \ref{fig:CNN}. Variants of CNN architectures used in different models discussed in section \ref{AppsMMDL} are VGGNet \cite{simonyan2014very}, GoogLeNet \cite{szegedy2015going}, ResNet \cite{he2016deep}, Inception \cite{szegedy2016rethinking}, Inception-ResNet-V2 \cite{szegedy2016inception}, ResNeXt101 \cite{xie2017aggregated}, Xception \cite{chollet2016xception}, Faster R-CNN \cite{ren2015faster}, Deep CNN , C3D \cite{tran2014learning}, and TextCNN \cite{gong2018does} architectures.

\begin{figure}
	\centering
	\tmpframe{\includegraphics[width= \textwidth, height=35mm]{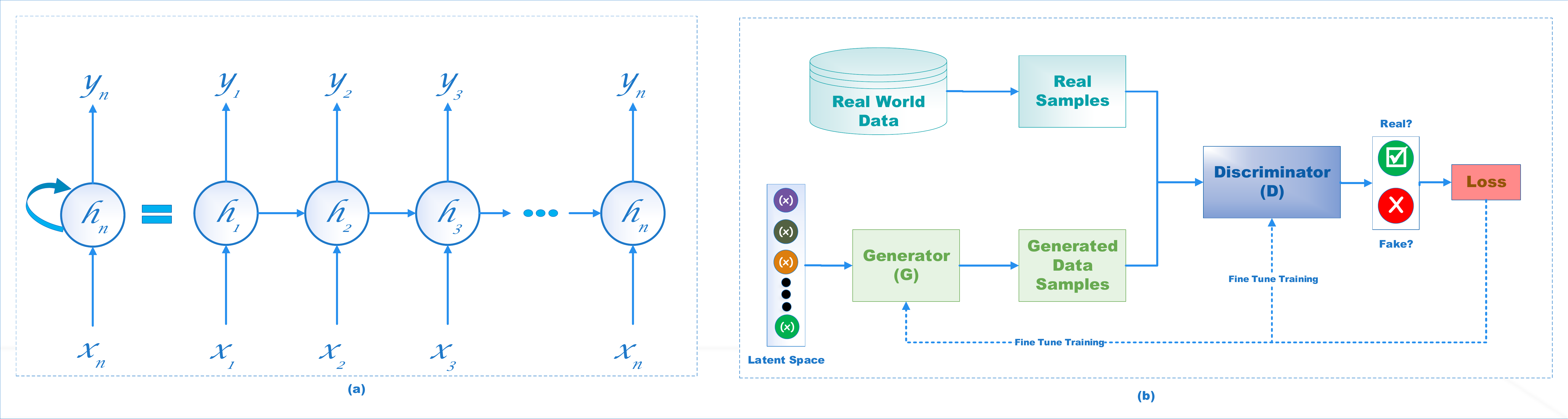}}
	\caption{General Diagram of (a) Recurrent Neural Network architecture and (b) Generative Adversarial Network architecture.}
	\label{fig:RNN}
\end{figure}

\textbf{Recurrent Neural Network (RNN)} architecture is used to enhance NN's capabilities, which can get inputs from fixed-length data to process variable-length sequences. RNN architecture can process one element at a time, and the output of hidden nodes is used as additional input for the next element. It can address speech synthesis, time series and NLP problems. A general diagram of Recurrent Neural Network architecture is shown in Figure \ref{fig:RNN}(a).
RNN architecture is further extended to variants of memory units, including LSTM \cite{hochreiter1997long} and GRU \cite{chung2014empirical}. Variants of RNN architectures used in different models discussed in section \ref{AppsMMDL} are Bidirectional RNN (BRNN) \cite{schuster1997bidirectional}, LSTM \cite{hochreiter1997long}, Bidirectional LSTM (BiLSTM) \cite{basaldella2018bidirectional}, GRU \cite{chung2014empirical}, and Stacked GRU (SGRU) \cite{wu2017cascade}.
J. Redmon et al. proposed a \textbf{You Only Look Once (YOLO)} architecture, which is efficiently used for real time object detection \cite{redmon2016you}. YOLO architecture is repurposed as a regression problem instead of a classification problem in this research. A single neural network is used in YOLO model to predict various object's bounding boxes and for the association of class probabilities. Object detection problem is more complex than classification problem because more objects are detected instead of one at a time, and the location of these objects is also detected. YOLO algorithm applies a clever CNN architecture to an entire image and then splits them into various regions. Then probabilities and bounding boxes of each region are predicted. The general architecture diagram of You Only Look Once is presented in Figure \ref{fig:YOLO}.

\subsection{Miscellaneous Architectures:}

Miscellaneous architectures like Support Vector Machine, Generative Adversarial Network and Hidden Markov model are used in the various model presented in section \ref{AppsMMDL}.
\textbf{Support Vector Machine (SVM)} is a traditional supervised learning approach of ML and is used to solve big data classification issues to help multi-domain applications \cite{suthaharan2016support}. SVM is characterized as a statistical learning algorithm to solve linear and non-linear regression or classification problems. The algorithm of SVM creates a hyperplane or line between data to separate it into classes. Its attempt to create a possibly wide separation between two classes that is helpful to make decisions. In this article, SVM and its variant SVM-RBF architectures are used in various models of retrieval-based and template-based image description tasks, emotion recognition tasks and event detection tasks. \textbf{Generative Adversarial Network (GAN)} is a hybrid architecture consisting of two primary components, i.e., generator and discriminator networks. These networks can be in the form of NN like ANN, CNN, RNN, LSTM etc. And discriminator network contains fully connected layers along with the classifier \cite{goodfellow2014generative}. GAN architecture consists of two NN compete to minimize the difference between actual input data and generated data. The discriminator network takes the real data and data generated from the generator network and attempts to identify whether data is real or fake. A general diagram of the Generative Adversarial Network is shown in Figure \ref{fig:RNN}(b). The generator network can produce new output data after training completion, i.e., not distinct from actual data. In this article, GAN and its variant WaveGAN architectures are used in various models of encoder-decoder based and attention-based image description tasks and speech synthesis tasks.
\textbf{Hidden Markov Model (HMM)} is a statistical model used first time in speech recognition tasks and biological sequences analysis \cite{franzese2019hidden}. These models are used to collect confidential information from sequential patterns. In HMM architecture, the modeling method is considered an unknown parametric Markov process, and the goal is to decide hidden parameters from given parameters. Many ML techniques are focused on successfully implementing HMMs in speech recognition, computational biology, character recognition, mobile communication techniques, bio-informatics, time series surveillance data, etc. In this article, HMM architectures are used in various template-based image descriptions, video descriptions, emotion detection and event detection tasks.

\begin{table}[]
	\centering
	\scriptsize
	\caption{Dataset splits of Flickr group.}
	\label{tab:Flickr}
	\begin{tabular}{lllllll}
		\cellcolor{blue!20} & \multicolumn{3}{c}{\cellcolor{blue!20}\textbf{Flickr8k}} & \multicolumn{3}{c}{\cellcolor{blue!13}\textbf{Flickr8k-CN}} \\
		& \textbf{Images} & \textbf{Caption/image} & \textbf{Total Captions} & \textbf{Images} & \textbf{Caption/image} & \textbf{Total Captions} \\
		
		\bottomrule
		\textbf{Training} & 6000 & 5 & 30,000 & 6000 & 5 & 30,000 \\
		\textbf{Validation} & 1000 & 5 & 5000 & 1000 & 5 & 5000 \\
		\textbf{Test} & 1000 & 5 & 5000 & 1000 & 5 & 5000 \\
		\bottomrule
		\textbf{Total} & 8000 & 5 & 40,000 & 8000 & 5 & 40,000 \\
		
		\cellcolor{blue!20}& \multicolumn{3}{c}{\cellcolor{blue!20}\textbf{Flickr30k}} & \multicolumn{3}{c}{\cellcolor{blue!13}\textbf{FlickrStyle10k}} \\
		\textbf{Training} & 29,783 & 5 & 1,48,915 & 7000 & 7 & 49,000 \\
		\textbf{Validation} & 1000 & 5 & 5000 & 2000 & 7 & 14,000 \\
		\textbf{Test} & 1000 & 5 & 5000 & 1000 & 10 & 10,000 \\
		\bottomrule
		\textbf{Total} & 31,783 & 5 & 158,915 & 10,000 & - & 73,000
	\end{tabular}
\end{table}

\section{MMDL Datasets:} 
\label{dataset}
A wide variety of datasets are available for multimodal deep learning applications. These datasets are a major driving force for accelerated development in the research area of various MMDL applications. In this review, the datasets used in multiple applications discussed in section \ref{AppsMMDL} are briefly described below.

\subsection{Datasets used for Image Description:}

In this literature, various datasets are used for image description task which are explained herewith, \textbf{Flickr} was first created in 2004 by Ludicorp. It is an American online community for hosting images and video services. Billions of High resolution as well as professional images, are available publicly in online community. Different versions of Flicker datasets are available for research i.e. Flickr-8k\cite{hodosh2013framing}, Flickr-8kCN \cite{li2016adding}, Flicker-30k \cite{young2014image}, and FlickrStyle10k \cite{gan2017stylenet}. A brief description of these datasets is given below, and their dataset splits are summarized in Table \ref{tab:Flickr}. \textbf{PASCAL dataset:} PASCAL Visual Object Classes (PASCAL-VOC) challenges from 2005 to 2012 provides datasets for object class recognition. Later, C. Rashtchian et al. \cite{rashtchian2010collecting} proposed a UIUC PASCAL sentence dataset used for image classification, object detection and object segmentation tasks. This dataset contains 1000 images, five human written descriptions in English language are generated for each image.

Another dataset for image description task is \textbf{SBU Captioned Photo}, that contains one million filtered images from Flickr with associated descriptions in the form of text. V. Ordonez et al. \cite{ordonez2011im2text} made the first attempt to prepare a large dataset for the image description task. The analysis shows some issues with SBU Captioned Photo dataset's utility, like captions containing irrelevant information (name, location of objects) compared to image content or brief descriptions for each image. And \textbf{Microsoft-Common Objects in Context (MS-COCO)} dataset contains 328,000 images used for object recognition, captioning, and segmentation.  \cite{lin2014microsoft}. MS-COCO dataset is categorized into 91 common objects categories, 82 out of 91 categories include more than 5000 label instances resulting in 2,500,000 label instances. This dataset contains more label instances as per category, and each image contains five descriptions. The dataset is split into a training set containing 164,000 images paired with five captions per image. And Validation and test sets comprise 82,000 images, each with five captions per image. MS-COCO Captions \cite{chen2015microsoft} and MS-COCO QA \cite{ren2015exploring} are other variants of this dataset, whose statistics are summarized in Table \ref{tab:COCO}.

\begin{table}[]
	\caption{MS-COCO dataset statistics }
	\label{tab:COCO}
	\scriptsize
	\begin{tabular}{p{1.1cm}p{0.7cm}p{1.1cm}p{2cm}p{1.8cm}p{0.7cm}p{1.1cm}p{2cm}p{1.8cm}}
		\cellcolor{blue!20}& \multicolumn{4}{c}{\cellcolor{blue!20}\textbf{MS-COCO}} & \multicolumn{4}{c}{\textbf{\cellcolor{blue!13}MS-COCO Caption}} \\
		& \textbf{Images} & \textbf{Categories} & \textbf{Label Instances} & \textbf{Captions/Image} & \textbf{Images} & \textbf{Categories} & \textbf{Label Instances} & \textbf{Captions/Image} \\
		\bottomrule
		\textbf{Training} & 164,000 & 91 & 2,500,000 & 5 & 166,000 & 80 & 1,500,000 & 5 \\
		\textbf{Validation} & 82,000 & 91 & 2,500,000 & 5 & 82,000 & 80 & 1,500,000 & 5 \\
		\textbf{Test} & 82,000 & 91 & 2,500,000 & 5 & 82,000 & 80 & 1,500,000 & 5 \\
		\bottomrule
		\textbf{Total} & 328,000 & 91 & 2,500,000 & 5 & 330,000 & 80 & 1,500,000 & 5
	\end{tabular}
\end{table}

\textbf{SentiCap} \cite{mathews2016senticap} dataset characterizes images with emotions by producing captions automatically with negative or positive sentiments. It is a subset of the MS-COCO dataset based on COCO images labeled with three negative and three positive sentiments. The positive subset includes 998 images with 2,873 sentences for training and 673 images with 2,019 sentences for testing; similarly, the negative subset contains 997 images with 2,468 sentences for training and 503 images with 1,509 sentences for testing. Statistics and splits of the SentiCap dataset are shown in Table \ref{tab:SentiCap}. 

% \usepackage{multirow}
\begin{table}[]
	\scriptsize
	\caption{SentiCap dataset statistics }
	\label{tab:SentiCap}
	\begin{tabular}{lllll}
		\bottomrule
		\cellcolor{blue!20}& \cellcolor{blue!20}\textbf{Dataset Splits} & \cellcolor{blue!20}\textbf{Number of Images} & \cellcolor{blue!20}\textbf{Number of Sentences} & \cellcolor{blue!20}\textbf{Number of Sentiments} \\
		\bottomrule
		\multicolumn{1}{c}{\multirow{2}{*}{\textbf{Positive Subset}}} & Training & 998 & 2,873 & 3 \\
		\multicolumn{1}{c}{} & Testing & 673 & 2,019 & 3 \\
		\bottomrule
		\multirow{2}{*}{\textbf{Negative Subset}} & Training & 997 & 2,468 & 3 \\
		& Testing & 503 & 1,509 & 3 \\
		\bottomrule
	\end{tabular}
\end{table}

\subsection{Datasets used for VQA task:}
In this section, datasets used for VQA task are discussed, like \textbf{Visual Question Answering (VQA) dataset} is the most frequently used dataset for the Visual Question Answering task. There are various variants of the VQA dataset like VQA v1.0 \cite{antol2015vqa}, VQA v2.0 \cite{wu2017visual}, VQA-CP \cite{agrawal2018don}, VQA-X \cite{huk2017attentive}, Ok-VQA \cite{marino2019ok}, and FVQA \cite{wang2017fvqa}. VQA v1.0 also named as VQA-real dataset is split into two parts; 1st part contains 123,287 training-validation images and 81,434 testing images from MS-COCO \cite{antol2015vqa}. Second part contains 50,000 abstract scenes which are used for high-level reasoning in VQA task. Ten answers for each question from different annotators is gathered in this version. 
VQA v2.0 \cite{wu2017visual} is proposed to remove the limitation of inherent bias in version 1.0 by adding another image with a different answer for the same question. Therefore, the VQA v2.0 dataset size is double as compared to version 1.0. VQA v2.0 dataset contains additional complimentary images to cope with the referred above limitation and hence contains approximately 443,000 (Question, Image) pairs for training, 214,000 pairs for validation, and 453,000 pairs for testing. Statistics of VQA v1.0 and 2.0 are summarized in Table \ref{tab:VQAv1nd2}.

% \usepackage{multirow}
\begin{table}[]
	\scriptsize
	\caption{VQA V1.0 and V2.0 dataset statistics }
	\label{tab:VQAv1nd2}
	\begin{tabular}{p{1.7cm}p{1cm}p{1.3cm}p{1.2cm}p{1.2cm}p{1.7cm}p{1.6cm}p{2.5cm}}
		\multicolumn{8}{c}{\cellcolor{blue!20}\textbf{VQA v1.0 dataset}} \\
		& \textbf{Dataset Splits} & \textbf{Number of Images} & \textbf{Questions/ Image} & \textbf{Answers/ Question} & \textbf{Questions Annotations} & \textbf{Answers Annotations} & \textbf{Complementary Pairs Annotations} \\
		\bottomrule
		\multirow{3}{*}{\parbox{2cm}{\textbf{VQA-Real Images}}} & Training & 82,783 & 3 & 10 & 248,349 & 2,483,490 & - \\
		& Validation & 40,504 & 3 & 10 & 121,512 & 1,215,120 & - \\
		& Testing & 81,434 & 3 & 10 & 244,302 & - & - \\
		\bottomrule
		\multirow{3}{*}{\parbox{2cm}{\textbf{VQA-Abstract Scenes}}} & Training & 20,000 & 3 & 10 & 60,000 & 600,000 & - \\
		& Validation & 10,000 & 3 & 10 & 30,000 & 300,000 & - \\
		& Testing & 20,000 & 3 & 10 & 60,000 & - & - \\
		\multicolumn{8}{c}{\cellcolor{blue!20}\textbf{VQA v2.0 dataset}} \\
		\multirow{3}{*}{\parbox{2cm}{\textbf{VQA-Balanced Real Images}}} & Training & 82,783 & - & 10 & 443,757 & 4,437,570 & 200,394 \\
		& Validation & 40,504 & - & 10 & 214,354 & 2,143,540 & 95,144 \\
		& Testing & 81,434 & - & 10 & 447,793 & - & - \\
		\bottomrule
		\multirow{2}{*}{\parbox{2cm}{\textbf{VQA-Balanced Abstract Scenes}}} & Training & 20,629 & - & 10 & 22,055 & 220,550 & - \\
		& Validation & 10,696 & - & 10 & 11,328 & 113,280 & - \\
		\bottomrule
	\end{tabular}
\end{table}

\textbf{DAtaset for QUestion Answering on Real-world images (DAQUAR)} \cite{malinowski2014multi} was the first significant dataset for the VQA task. DAQUAR dataset contains 6,795 QA pairs for the training set and 5,673 QA pairs for the testing set on 1,449 images from the NYU-DepthV2 dataset. \textbf{Visual7W} \cite{zhu2016visual7w} is another dataset used for the VQA task that contains object localization and dense annotations in an image. Visual7W dataset consists of seven questions (what, when, where, why, who, which, and how.) The questions in Visual7W are much richer, and their answers are also longer than the VQA dataset. It consists of 327,939 QA pairs on 47,300 images from the COCO dataset. 
\textbf{Visual Genome} dataset is a huge structured knowledge presentation of visual perceptions and a complete collection of question-answers and descriptions \cite{krishna2017visual}. Dense annotations of attributes, relationships, and objects within each image are collected in this dataset. This dataset contains approximately 108,000 images.

\textbf{Compositional Language and Elementary Visual Reasoning diagnostics (CLEVR)} dataset \cite{johnson2017clevr} is used to solve complex reasoning. It performs extensive diagnostic for better understanding of reasoning capabilities. It contains 100,000 images, out of which 70,000, 15,000, and 15,000 images are used for training, testing, and validation sets, respectively. 
\textbf{Task Directed Image Understanding Challenge (TDIUC)} dataset \cite{kafle2017analysis} contains 167,437 Images from Visual Genome and MS-COCO datasets. TDIUC dataset contains 1,654,167 QA pairs derived from three different sources, which are organized into 12 separate categories. Statistics of other VQA datasets are summarized in Table \ref{tab:VQA}.

\begin{table}[]
	\scriptsize
	\caption{Statistics of different Visual Question Answering datasets.}
	\label{tab:VQA}
	\begin{tabular}{lllll}
		\bottomrule
		\cellcolor{blue!20}\textbf{Dataset Name} & \cellcolor{blue!20}\textbf{Number of Images} & \cellcolor{blue!20}\textbf{Number of Questions} & \cellcolor{blue!20}\textbf{Questions Type} & \cellcolor{blue!20}\textbf{Unique Answers} \\
		\bottomrule
		DAQUAR & 1,449 & 16,590 & 3 & 968 \\
		COCO-QA & 123,287 & 117,684 & 4 & 430 \\
		Visual7W & 47,300 & 327,939 & 7 & 25,553 \\
		Visual Genome & 108,000 & 1,773,358 & 6 & 207,675 \\
		TDIUC & 167,437 & 1,654,167 & 12 & 1,618\\
		\bottomrule
	\end{tabular}
\end{table}

\subsection{Datasets used for Video Description:}

In this section, dataset used for video description task are explained briefly, like \textbf{TACoS-MultiLevel} dataset
\cite{rohrbach2014coherent} is also annotated on TAcos Corpus \cite{regneri2013grounding} through AMT workers. TACoS-MultiLevel dataset contains 185 long indoor videos, and approximately every video is six minutes long. These videos contain various actors, small interacting objects, and activities about cooking scenarios. Multiple AMT workers annotate the video sequence intervals by pairing them with a short sentence. Statistics of the various dataset used for video description task is summarized in Table \ref{tab:VidDescription}.
And \textbf{Microsoft Video Description (MSVD)} dataset \cite{chen2011collecting} corpus contains a different crowd-sourced description of small video clips in the form of text. MSVD dataset contains 1,970 clips from YouTube annotated in the format of sentences by AMT workers. Approximately, 10-25 seconds duration video clip is used to illustrate activity in this dataset, and it also supports multilingual descriptions like in the form of English, Chinese, German, etc. Similarly \textbf{Montreal-Video Annotation Dataset (M-VAD)} \cite{torabi2015using} consists of 48,986 video clips from ninety-two different movies based on descriptive video service. M-VAD dataset contains 92 filtered and 92 unfiltered movies. On Average, each video clip is spanned over 6.2 seconds and the entire running time for the whole dataset is approximately 84.6 hours. It contains 510,933 words, 48,986 paragraphs and 55,904 sentences. The dataset is split into 38,949 clips for training, 4,888 clips for validation, and 5,149 clips for testing.

\textbf{Microsoft Research-Video to Text(MSR-VTT)} dataset \cite{xu2016msr} consists of a broader range of open domain videos used for videos' description task. MSR-VTT contains approximately a total of 7,180 videos, which are sub-divided into 10,000 video clips. Twenty different categories are organized to group these clips. The dataset is split into a training, validation, and testing set containing 6,513, 497 and 2,990 videos. Twenty captions are annotated for each video by AMT workers. 
\textbf{Charades} dataset \cite{sigurdsson2016hollywood} includes 9,848 videos of various daily life indoor activities. Different 267 AMT workers are deployed to captures these activities in the form of videos from three continents. The recordings are captured based on a given script that contains descriptions of actions and objects to be recorded. Recording of these videos is accomplished through 15 different indoor scenarios. In this dataset, 157 action classes and 46 objects are used for the description of scenes. It provides 27,847 descriptions for all captured indoor scenes. 
\textbf{ActivityNet Captions} dataset \cite{krishna2017dense} includes 100,000 dense descriptions in natural language for approximately 20,000 videos from ActivityNet \cite{caba2015activitynet}. The total period for these videos is about 849 hours. In this article, ActivityNet dataset is used in video description task. 
And \textbf{Video-to-Commonsense (V2C)} dataset \cite{fang2020video2commonsense} consists of approximately nine thousand human agents videos implementing different actions from MSR-VTT dataset. Three kinds of commonsense descriptions are used to annotate these videos by recruiting two sets of AMT workers. 6,819 videos are used for the training set, and 2,906 are used for the testing set. Approximately 121,65 commonsense captions are generated for these videos. Extension of this dataset named V2C-QA is also proposed to deal with the VQA task in the perspective of commonsense knowledge. 

\begin{table}[]
	\centering
	\scriptsize
	\caption{Statistics of video description method datasets.}
	\label{tab:VidDescription}
	\begin{tabular}{p{3.9cm}p{1.7cm}p{1.5cm}p{1.5cm}p{1.7cm}p{1.5cm}p{1.5cm}}
		\bottomrule
		\cellcolor{blue!20}\textbf{Dataset Name} & \cellcolor{blue!20}\textbf{Number of Classes} & \cellcolor{blue!20}\textbf{Total Videos} & \cellcolor{blue!20}\textbf{Number of clips} & \cellcolor{blue!20}\textbf{Number of sentences} & \cellcolor{blue!20}\textbf{Number of Words} & \cellcolor{blue!20}\textbf{Length (Hours)} \\
		\bottomrule
		TACoS-MultiLevel \cite{rohrbach2014coherent}& 1 & 185 & 14,105 & 52,593 & 2,000 & 27.1 \\
		MSVD \cite{chen2011collecting}& 218 & 1,970 & 1,970 & 70,028 & 607,339 & 5.3 \\
		M-VAD \cite{torabi2015using}& - & 92 & 48,986 & 55,904 & 519,933 & 84.6 \\
		MSR-VTT \cite{xu2016msr}& 20 & 7,180 & 10,000 & 200,000 & 1,856,523 & 41.2 \\
		Charades \cite{sigurdsson2016hollywood}& 157 & 9,848 & - & 27,847 & - & 82.01 \\
		ActivityNet Captions \cite{krishna2017dense}& - & 20,000 & - & 100,000 & 1,348,000 & 849.0\\
		\bottomrule
	\end{tabular}
\end{table}

\subsection{Datasets used for Speech Synthesis:}
Datasets used for speech synthesis task are explained briefly herewith, like
\textbf{North American English} dataset \cite{wang2017tacotron} are built using a single speaker speech database. The duration of speech data in this dataset is around 24.6 hours. Professional female speakers are used to speak this speech data. This dataset is used for the speech synthesis process.
\textbf{Voice Cloning Toolkit (VCTK)} dataset \cite{veaux2016superseded} contains read speech data from 109 native English speakers with different accents. Speakers read approximately 400 sentences chosen from the newspaper. The cumulative length of these audios is about 44 hours, with a sample rate of 48-kHz. Standard train and test splits are not provided in this dataset. 
\textbf{LibriSpeech} corpus \cite{panayotov2015librispeech} contains English read speech from audiobooks that are derived through the LibriVox project. Around 2484 male and female speakers are engaged to create this corpus. The cumulative length of these audios is approximately 1000 hours with a sample rate of 16-kHz.

\textbf{Lj Speech} dataset \cite{ito2017lj} is a publicly available speech dataset; it contains 13,100 small audio clips. These clips are recorded by a single speaker, who read passages from 7 different books.  The duration of each clip is different, can vary from 1 to 10 seconds. The cumulative period of these clips is around 24 hours. 
And \textbf{Proprietary Speech} \cite{elias2020parallel} dataset is used for the speech synthesis process. The cumulative duration of speech data in this dataset is around 405 hours. It includes a total of 347,872 utterances varying in three different English accents from 45 speakers. Out of these 45 speakers, 32 speakers have a US English accent, 8 have a British English accent, and 5 have an Australian English accent.

\subsection{Datasets used for Emotion Recognition:}

\begin{table}[]
	\scriptsize
	\caption{Statistics of different emotion detection databases.}
	\label{tab:EmotionDetection}
	\begin{tabular}{lllll}
		\bottomrule
		\cellcolor{blue!20}\textbf{Dataset Name} & \cellcolor{blue!20}\textbf{Language} & \cellcolor{blue!20}\textbf{Number of subjects} & \cellcolor{blue!20}\textbf{Number of samples} & \cellcolor{blue!20}\textbf{Emotion states} \\
		\bottomrule
		eNTERFACE \cite{martin2006enterface}& English & 42 (34-male, 8-female) & 1166 video sequences & 6 \\
		FABO \cite{gunes2006bimodal}& English & 23 (11-male, 12-female) & Camera recording of 1 hour/subject & 10 \\
		IEMOCAP \cite{busso2008iemocap}& English & 10 (5-male, 5-female) & 12 hours & 5 \\
		MSP-Improve \cite{busso2016msp}& English & 12 (6-male, 6-female) & 18 hours & 4 \\
		MuSE \cite{jaiswal2019muse}& English & 28 (19-male, 9-female) & 10 hours & 2 \\ 
		SEMAINE \cite{mckeown2011semaine}& English & 150 (57-male,93-female) & 959 conversations & 5 \\
		RECOLA \cite{ringeval2013introducing}& French& 46 (19-male, 27-female)& 7 hours& 5 \\
		\bottomrule
	\end{tabular}
\end{table}

In this section, datasets used for emotion recognition task are explained briefly. \textbf{eNTERFACE audio-visual} database \cite{martin2006enterface} contains 1,166 video sequences. Out of these sequences, 902 covers male video recordings, and 264 covers female recordings. Six different emotion categories are expressed in this database, i.e., happiness, anger, fear, disgust, surprise, and sadness. Statistics of different datasets used for emotion detection task is summarized in Table \ref{tab:EmotionDetection}.
Hatice Gunes and Massimo Piccardi produced the \textbf{Bimodal Face and Body Gesture Database (FABO)} database \cite{gunes2006bimodal} at the University of Technology, Sydney. This database creates bimodal face and body expressions for the automated study of human affective activities. Various volunteers gathered visual data in a laboratory environment by directing and requesting the participants with the perspective of desired actions/movements. Ten different emotion categories are expressed in this database, i.e., neutral, happiness, anger, fear, disgust, surprise, sadness, uncertainty, anxiety, and boredom.
\textbf{Interactive Emotional Dyadic Motion Capture (IEMOCAP)} \cite{busso2008iemocap} is a multi-speaker and multimodal database. The cumulative duration of audio-visual data in this dataset is around 12 hours. Various videos, face motions, speech, text transcriptions are included in this bunch of data. Multiple annotators are used to annotating this dataset into categorical labels like happiness, frustration, anger, sadness, neutrality, and dimensional labels like activation, valence, and dominance. 

\textbf{MSP-Improve} \cite{busso2016msp} database was compiled to record naturalistic emotions from improvised circumstances. Six sessions are recorded from 12 actors, and 652 target sentences are collected containing lexical data. The duration of each session is around three hours. Data is split into 2,785 natural interactions, 4,381 improvised turns, and 620 read sentences for 8,438 utterances in total.
\textbf{Multimodal Stressed Emotion (MuSE)} \cite{jaiswal2019muse} database recognizes the interaction between emotion and stress in naturally spoken communication. This dataset contains 55 different recordings from 28 contestants. Each contestant was recorded for stressed and not-stressed sessions. The cumulative duration of this dataset is approximately 10 hours from 2,648 utterances. 
\textbf{Sustained Emotionally coloured Machinehuman Interaction using Nonverbal Expression (SEMAINE)} \cite{mckeown2011semaine} has built a substantial audio-visual database to create Sensitive Artificial Listener agents that involve a person in a prolonged emotional conversation. Five high-resolution cameras and four microphones are used synchronously for high-quality recording. One hundred fifty contestants from eight different countries record 959 conversations in total. \textbf{REmote COLlaborative and Affective (RECOLA)} database \cite{ringeval2013introducing} consists of spontaneous and natural emotions in continuous domain. Video, audio, electrocardiogram, and electro-dermal activity modalities are used in this database. The cumulative duration of the recording is around 9.5 hours from 46 French-speaking contestants.

\section{MMDL Evaluation Metrics:}
\label{E-Metrics}
Evaluation metrics determine the performance of machine learning or statistical models. For any project, evaluation of algorithms and ML models are essential. There are several kinds of evaluation metrics available to validate the frameworks. To test the model, single as well as multiple evaluation metrics are used. Because the performance of architecture may vary to different evaluation metrics, it may show better results using one metric but shows different results using other metrics. Therefore, evaluating the performance of a model with the perspective of many evaluation metrics gives a broader view. Overview of Various Evaluation metrics used in this article is summarized and presented with its relevant dataset and MMDL application group in Table \ref{tab:EvaluationMetrics}.

\begin{table}[]
	\scriptsize 
	\centering
	\caption{Overview of Evaluation metrics used in MMDL Applications. }
	\label{tab:EvaluationMetrics}
	\begin{adjustbox}{width=1.1\textwidth, center=\textwidth}
		\begin{tabular}{p{5cm}p{7.5cm}p{4.4cm}}
			\bottomrule
			\cellcolor{blue!20}\textbf{Evaluation Metrics} & \cellcolor{blue!20}\textbf{Supporting Datasets} & \cellcolor{blue!20}\textbf{MMDL Application Group} \\
			\bottomrule
			BLEU \cite{wu2017cascade,guo2019mscap,he2019image,feng2019unsupervised,wang2018novel,cao2019image,cheng2020stack,li2017gla,anderson2018bottom,liu2020chinese,liu2020image,wang2020cross,wei2020multi,jiasen2020adaptive,krishna2017dense,wang2018video,pei2019memory,aafaq2019video,liu2020sibnet,yu2016video,fang2020video2commonsense,rahman2020semantically,wang2018video,chen2018less,li2019end,mun2019streamlined,zhang2019reconstruct,xu2020deep,wei2020exploiting,patro2020robust} & Pascal Sentence, Flickr8k, Flickr30k, FlickrStyle10k, MS-COCO, MS-COCO Caption, SentiCap, Visual Genome, VQA2.0, TACoS-MultiLevel, ActivityNet Captions, MSVD, MSR-VTT, V2C, Charades, VQA-X & Image Description, Video Description, VQA \\
			ROUGE \cite{feng2019unsupervised,cheng2020stack,li2017gla,anderson2018bottom,liu2020chinese,liu2020image,wang2020cross,wei2020multi,jiasen2020adaptive,wang2018reconstruction,pei2019memory,aafaq2019video,liu2020sibnet,fang2020video2commonsense,wang2018video,chen2018less,li2019end,zhang2019reconstruct,xu2020deep,patro2020robust} & Pascal Sentence, Flickr8k, Flickr8k-CN, Flickr30k, MS-COCO, Visual Genome, VQA2.0, MSVD, MSR-VTT, V2C, Charades, VQA-X & Image Description , Video Description, VQA \\
			PPLX \cite{guo2019mscap} & MS-COCO, Pascal Sentence, Flickr8k, FlickrStyle10k, SentiCap & Image Description, Video Description \\
			METEOR \cite{guo2019mscap,he2019image,feng2019unsupervised,wang2018novel,cao2019image,cheng2020stack,li2017gla,anderson2018bottom,liu2020chinese,liu2020image,wang2020cross,wei2020multi,jiasen2020adaptive,krishna2017dense,wang2018reconstruction,pei2019memory,aafaq2019spatio,liu2020sibnet,yu2016video,fang2020video2commonsense,wang2018video,chen2018less,li2019end,mun2019streamlined,zhang2019reconstruct,xu2020deep,wei2020exploiting,patro2020robust} & MS-COCO, Pascal Sentence, Flickr8k, Flickr30k, FlickrStyle10k, SentiCap, Visual Genome, VQA2.0, MSVD, M-VAD, ActivityNet Captions, MSR-VTT, TACoS-MultiLevel, V2C, Charades, VQA-X & Image Description, Video Description, VQA \\
			CIDEr \cite{wu2017cascade,guo2019mscap,he2019image,feng2019unsupervised,wang2018novel,cao2019image,cheng2020stack,li2017gla,anderson2018bottom,liu2020chinese,liu2020image,wang2020cross,wei2020multi,jiasen2020adaptive,krishna2017dense,wang2018reconstruction,aafaq2019spatio,liu2020sibnet,yu2016video,wang2018video,chen2018less,li2019end,mun2019streamlined,zhang2019reconstruct,xu2020deep,wei2020exploiting,patro2020robust} & MS-COCO, Flickr8k, Flickr8k-CN, Flickr30k, FlickrStyle10k, SentiCap, Visual Genome, VQA2.0, ActivityNet Captions, MSVD, MSR-VTT, TACoS-MultiLevel, Charades, VQA-X & Image Description, Video Description, VQA \\
			SPICE \cite{feng2019unsupervised,anderson2018bottom,wei2020multi,patro2020robust}& MS-COCO, Visual Genome, VQA2.0, VQA-X & Image Description, VQA \\
			WUPS \cite{ren2015exploring,wang2017vqa,xi2020visual,wang2018fvqa}& DAQUAR, COCO-QA, VQA, FVQA, Visual Genome & VQA \\
			Accuracy \cite{ben2017mutan,desta2018object,cadene2019murel,lobry2020rsvqa,wang2017vqa,yu2017multi,li2019relation,xi2020visual,guo2020re,wang2018fvqa,narasimhan2018straight,yu2020cross,basu2020aqua,arik2018neural,huang2017fusion,nguyen2017deep,nguyen2018deep,tripathi2018multi,hazarika2018icon,jaiswal2019controlling,chong2019emochat,li2020multistep,huan2020video,cimtay2020cross,gwon2016multimodal,huang2018learning} & DAQUAR, COCO-QA, VQA, VQA2.0, VQA-CP, FVQA, OK-VQA, Visual Genome, Visual7W, CLEVR, TDIUC, LibriSpeech, VCTK, eNTERFACE, FABO, IEMOCAP, SEMAINE, MuSE, MSP-Improve, RECOLA, MED & VQA, Speech Synthesis, Emotion Recognition, Event Detection \\
			TopK \cite{hu2017modeling,fang2020video2commonsense,wang2018fvqa,narasimhan2018straight,yu2020cross,taigman2018VoiceLoopVF}& Visual Genome, V2C-QA, FVQA, OK-VQA & Image Captioning, VQA \\
			MOS \cite{wang2017tacotron,arik2017deep,gibiansky2017deep,ping2017deep,oord2018parallel,arik2018neural,taigman2018VoiceLoopVF,shen2018natural,elias2020parallel} & North American English, MagnaTagATune, VCTK, LibriSpeech, LJ Speech & Speech Synthesis \\
			EER \cite{arik2018neural,huang2018learning}& LibriSpeech, VCTK, UCSD & Speech Synthesis, Event Detection \\
			MCD \cite{taigman2018VoiceLoopVF}& LJ Speech, VCTK & Speech Synthesis \\
			F1-Score \cite{hazarika2018icon,hazarika2018self,li2020multistep,lai2020different,huan2020video,gao2017event,yang2019shared}& IEMOCAP, SEMAINE, RECOLA, MED & Emotion Recognition, Event Detection \\
			MAE \cite{hazarika2018self}& IEMOCAP, SEMAINE & Emotion Recognition \\
			Unweighted Average Recall \cite{li2020multistep} & IEMOCAP, RECOLA & Emotion Recognition \\
			mAP \cite{gwon2016multimodal} & TRECVID, MED & Event Detection \\
			AUC \cite{huang2018learning,koutras2018exploring}& UCSD, Avenue, COGNIMUSE & Event Detection \\
			EDR \cite{huang2018learning}& UCSD, Avenue & Event Detection \\
			\bottomrule
		\end{tabular}
	\end{adjustbox}
\end{table}

\section{Discussion and Future Direction:}
\label{Disc&future}
Issues and limitations or various applications mentioned in section \ref{AppsMMDL} are highlighted below. Their’s possible future research directions are also proposed separately. 

\subsubsection{Multimodal Image Description:}

In recent years, the image description domain improves a lot, but there is still some space available for improvement in some areas. Some methods did not detect prominent attributes, objects and, to some extent does not generate related or multiple captions. Here, some issues and their possible future directions are mentioned below to make some advancement in the image description.

\begin{itemize}
	\item The produced captions' accuracy depends mostly on grammatically correct captions, which depends on a powerful language generation model. Image captioning approaches strongly depend on the quality and quantity of the training data. Therefore, another key challenge in the future will be to create large-scale image description datasets with accurate and detailed text descriptions in an appropriate manner.
	\item Exiting models demonstrate their output on datasets comprising a collection of images from the same domain. Therefore, implement and test image description models on open-source datasets will be another significant research perspective in the future.
	\item Too much work has been done on supervised learning approaches, and training requires a vast volume of labeled data. Consequently, reinforcement and unsupervised learning approaches have less amount of work done and have a lot of space for improvement in the future.
	\item More advanced attention-based image captioning mechanism or captioning based on regions/multi-region is also a research direction in the future.
\end{itemize}

\subsubsection{Multimodal Video Description:}

The performance of the video description process is enhanced with the advancement of DL techniques. Even though the present video description method’s performance is already far underneath the human description process, this accuracy rate is decreasing steadily, and still, there is sufficient space available for enhancement in this area. Here, some issues and their possible future directions are mentioned below to make some advancement in the video description.

\begin{itemize}
	\item It is expected in the future that humans would be capable of interacting with robots as humans can interact with each other. Video dialogue is a promising area to cope with this circumstance similarly to audio dialogue (like, Hello Google, Siri, ECHO, and Alexa). In visual Dialogue, the system needs to be trained to communicate with humans or robots in a conversation/dialogue manner by ignoring conversational statements' correctness and wrongness. 
	\item In CV, most research has focused on descriptions of visual contents without extraction of audio/speech features from the video. Existing approaches extract features from visual frames or clips for a description of the video. Therefore, extraction of these audio features may be more beneficial for the improvement of this description process. Audio features, like the sound of water splash, cat, dog, car, guitar, etc., provide occasional/eventual information when no visual clue is available for their existence in the video.
	\item Existing approaches have performed end-to-end training, resultantly more and more data is utilized to improve the accuracy of the model. But extensive dataset still does not cover the occurrences of real-world events. To improve system performance in the future, learning from data itself is a better choice and achieves its optimum computational power. 
	\item In the video description process, mostly humans describe the visual content based on extensive prior knowledge. They do not all the time rely solely on visual contents; some additional background knowledge is applied by the domain expertise as well. Augmentation of video description approaches with some existing external knowledge would be an attractive and promising technique in this research area. This technique shows some better performance in VQA. Most likely, this would dominate some accuracy improvement in this domain as well.
	\item Video description can be used in combination with machine translation for auto video subtitling. This combination is very beneficial for entertainment and other areas as well. So this combination needs to be focused on making the video subtitling process easier and cost-effective in the future. 
	
\end{itemize}

\subsubsection{Multimodal Visual Question Answering:} 

In recent years, the VQA task has gained tremendous attention and accelerated the development process by the contribution of various authors in this specific domain. VQA is especially enticing because it comprises a complete AI task that considers free-form, open-ended answers to questions by extracting features from an image/video and the question asked. The accuracy of VQA research still goes beyond answering the visual-based questions as compared to human equivalent accuracy. Some issues and their possible future directions for the VQA task are mentioned below.

\begin{itemize}
	\item So far, the image/video feature extraction part is almost fixed to a model like, most of the time ImageNet Challenge model is used. In this model, image features  are extracted by the split of frames into uniform boxes that works well for object detection, but VQA needs features by tracking all objects using semantic segmentation. Therefore, some more visual feature extraction mechanisms for VQA tasks need to be explored in the future. 
	\item Goal-oriented datasets for VQA research need to be designed in feature to support real-time applications like instructing users to play the game, helping visually impaired peoples, support for data extraction from colossal data pool and robot interaction, etc. So far, VizWiz \cite{gurari2018vizwiz} and VQA-MED \cite{abacha2019vqa}are two publicly available goal-oriented datasets for VQA research. Therefore, in the future, more goal-oriented datasets need to be built for mentioned above applications. 
	\item Most baseline VQA methods have been tested using traditional accuracy measure that is adequate for the multiple-choice format. In the future, the assessment or evaluation method for open-ended frameworks for VQA tasks is examined to improve these models' accuracy. 
\end{itemize}

\subsubsection{Multimodal Speech synthesis:}
The DLTTS models use distributed representations and complete context information to substitute the clustering phase of the decision tree in HMM models. Therefore, to produce a better quality of speech synthesis process compared to the traditional method, several hidden layers are used to map context features to high dimensional acoustic features by DLTTS methods. More hidden layers inevitably raise the number of system parameters to achieve better performance. As a result, space and time complexity for system training is also increased. Therefore for network training, a large amount of computational resources and corpora is required. Besides these achievements, there is still room available for DLTTS models in terms of quality improvement in intelligibility and speech's naturalness. Therefore, some issues and their feature research directions are discussed below.

\begin{itemize}
	\item DLTTS approaches usually require a huge amount of high-quality (text, speech) pairs for training, which is time-consuming and an expensive process. Therefore in the future, data efficiency for E2E DLTTS  models training is improved by the publicly availability of unpaired text and speech recordings on a large scale.
	\item Little progress is made in front-end text analysis to extract valuable context or linguistic features to minimize the gap between the text-speech synthesis process. Therefore it is a good direction in the future to utilize specific context or linguistic information for E2E DLTTS systems.
	\item Parallelization will be an essential aspect for DLTTS systems to improve system efficiency because most DNN architecture needs many calculations. Some frameworks proposed in recent years also use parallel networks for training or inference and achieve some good results, but there is still room available to achieve optimum results in the future.  
	\item As for DLTTS application concerns, the use of speech synthesis for other real-world applications like voice conversion or translation, cross-lingual speech conversion, audio-video speech synthesis, etc., are good future research directions. 
\end{itemize}

\subsubsection{Multimodal Emotion Recognition:}

Analysis of automatic emotions requires some advanced modeling and recognition techniques along with AI systems. For more advanced emotion recognition systems future research based on AI based automated systems constitutes more scientific progress in this area. Some future directions of listed issues of multimodal emotion recognition are mentioned below.

\begin{itemize}
	\item Existing baseline approaches are successful, but further experience, knowledge, and tools regarding the analysis and measurement of automatic non-invasive emotions are required. 
	\item Humans use more modalities to recognize emotions and are compatible with signal processing; machines are expected to behave similarly. But the performance of automated systems is limited with a restricted set of data. Therefore, to overcome this limitation, new multimodal recordings with a more representative collection of subjects are needed to be considered in the future.
	\item The preprocessing complexity of physiological signals in emotion detection is a big challenge. In physiological signals, detection of emotion states through electrocardiogram, electromyography, and skin temperature is still emerging. Hence, to determine the potency of these techniques a detailed research can be carried out in the future.
\end{itemize}

\subsubsection{Multimodal Event Detection:}

Due to the exponential increase in web data, multimodal event detection has attracted significant research attention in recent years. MMED seeks to determine a collection of real-world events in a large set of social media data. Subspace learning is an efficient approach to handle the issue of heterogeneity for the learning of features from multimodal data. On the other hand, E2E learning models are more versatile because they are structured to obtain heterogeneous data correlations directly. MMED from social data is still challenging and needs to be improved in some open issues in the future.

\begin{itemize}
	\item The "curse of dimensionality" issue is raised with the concatenation of various feature vectors of multiple modalities. Some considerable advancement has been made in curse of dimensionality issue, but existing techniques still require some improvement in terms of achieving better learning accuracy for social platforms. Like, GAN or some RNN architectures' extensions are valuable to improve feature learning accuracy of multimodal social data.
	\item Textual data also coincide with audio and video media. Therefore, MMED is improved if this textual information is considered jointly with these media.
	\item Current research primarily focuses on the identification of events from a single social platform. A comprehensive understanding of social data events is obtained by synthesizing the information from multiple social platforms. Hence, event detection methods can be implemented simultaneously to examine social data from multiple platforms using a transfer learning strategy.
	
\end{itemize}

\section{Conclusion:}
\label{conclusion}
In this survey, we discussed the recent advancements and trends in multimodal deep learning. Various DL models are categorized into different application groups and explained thoroughly using multiple media. This article focuses on an up-to-date review of numerous applications using various modalities like image, audio, video, text, body gesture, facial expression, and physiological signals compared to previous related surveys. A novel fine-grained taxonomy of various multimodal DL applications is proposed. Additionally, a brief discussion on architectures, datasets, and evaluation metrics used in these MMDL applications is provided. Finally, open research problems are also mentioned separately for each group of applications, and possible research directions for the future are listed in detail. We expected that our proposed taxonomy and research directions would promote future studies in multimodal deep learning and helps in a better understanding of unresolved issues in this particular area.

%
% The acknowledgments section is defined using the "acks" environment (and NOT an unnumbered section). This ensures
% the proper identification of the section in the article metadata, and the consistent spelling of the heading.
\begin{acks}
I am thankful to my supervisor Prof Xi Li. I am incredibly pleased for the time he has invested in teaching everything this survey required in resolving issues that arose and for being such a good mentor.
\end{acks}

%
% The next two lines define the bibliography style to be used, and the bibliography file.

\bibliographystyle{ACM-Reference-Format}
\bibliography{ref}

% 
% If your work has an appendix, this is the place to put it.

\appendix

\end{document}